\documentclass[letterpaper]{article} % DO NOT CHANGE THIS
\usepackage{aaai25}  % DO NOT CHANGE THIS
\usepackage{times}  % DO NOT CHANGE THIS
\usepackage{helvet}  % DO NOT CHANGE THIS
\usepackage{courier}  % DO NOT CHANGE THIS
\usepackage[hyphens]{url}  % DO NOT CHANGE THIS
\usepackage{graphicx} % DO NOT CHANGE THIS
\usepackage{natbib}  % DO NOT CHANGE THIS AND DO NOT ADD ANY OPTIONS TO IT
\usepackage{caption} % DO NOT CHANGE THIS AND DO NOT ADD ANY OPTIONS TO IT
\usepackage{algorithm}
\usepackage{algorithmic}
\usepackage{amsmath}
\usepackage{amssymb}
\usepackage{booktabs}
\usepackage{array}
\usepackage{bm}
\usepackage[final]{pdfpages}

\usepackage{newfloat}
\usepackage{listings}
\DeclareCaptionStyle{ruled}{labelfont=normalfont,labelsep=colon,strut=off} % DO NOT CHANGE THIS
\floatstyle{ruled}
\newfloat{listing}{tb}{lst}{}
\floatname{listing}{Listing}
\title{Sparis: Neural Implicit Surface Reconstruction of Indoor Scenes \\ from Sparse Views}
\author{
    Yulun Wu\textsuperscript{\rm 1, \rm 2}\equalcontrib, Han Huang\textsuperscript{\rm 1, \rm 2}\equalcontrib, Wenyuan Zhang\textsuperscript{\rm 1, \rm 2}, Chao Deng\textsuperscript{\rm 1, \rm 2},\\
    Ge Gao\textsuperscript{\rm 1, \rm 2}\thanks{Corresponding author.}, Ming Gu\textsuperscript{\rm 1, \rm 2}, Yu-Shen Liu\textsuperscript{\rm 2}\\
}
\affiliations{
    \textsuperscript{\rm 1}Beijing National Research Center for Information Science and Technology (BNRist), Tsinghua University, Beijing, China\\
    \textsuperscript{\rm 2}School of Software, Tsinghua University, Beijing, China
    \{wu-yl22, h-huang20, zhangwen21, dengc23\}@mails.tsinghua.edu.cn, \{gaoge, guming, liuyushen\}@tsinghua.edu.cn
}

\usepackage{bibentry}
\begin{document}

\maketitle

\begin{abstract}
In recent years, reconstructing indoor scene geometry from multi-view images has achieved encouraging accomplishments. Current methods incorporate monocular priors into neural implicit surface models to achieve high-quality reconstructions. However, these methods require hundreds of images for scene reconstruction. When only a limited number of views are available as input, the performance of monocular priors deteriorates due to scale ambiguity, leading to the collapse of the reconstructed scene geometry. In this paper, we propose a new method, named \textit{Sparis}, for indoor surface reconstruction from sparse views. Specifically, we investigate the impact of monocular priors on sparse scene reconstruction, introducing a novel prior based on inter-image matching information. Our prior offers more accurate depth information while ensuring cross-view matching consistency. Additionally, we employ an angular filter strategy and an epipolar matching weight function, aiming to reduce errors due to view matching inaccuracies, thereby refining the inter-image prior for improved reconstruction accuracy. The experiments conducted on widely used benchmarks demonstrate superior performance in sparse-view scene reconstruction.
\end{abstract}


\begin{figure*}[ht]
    \centering
    \includegraphics[width=\textwidth]{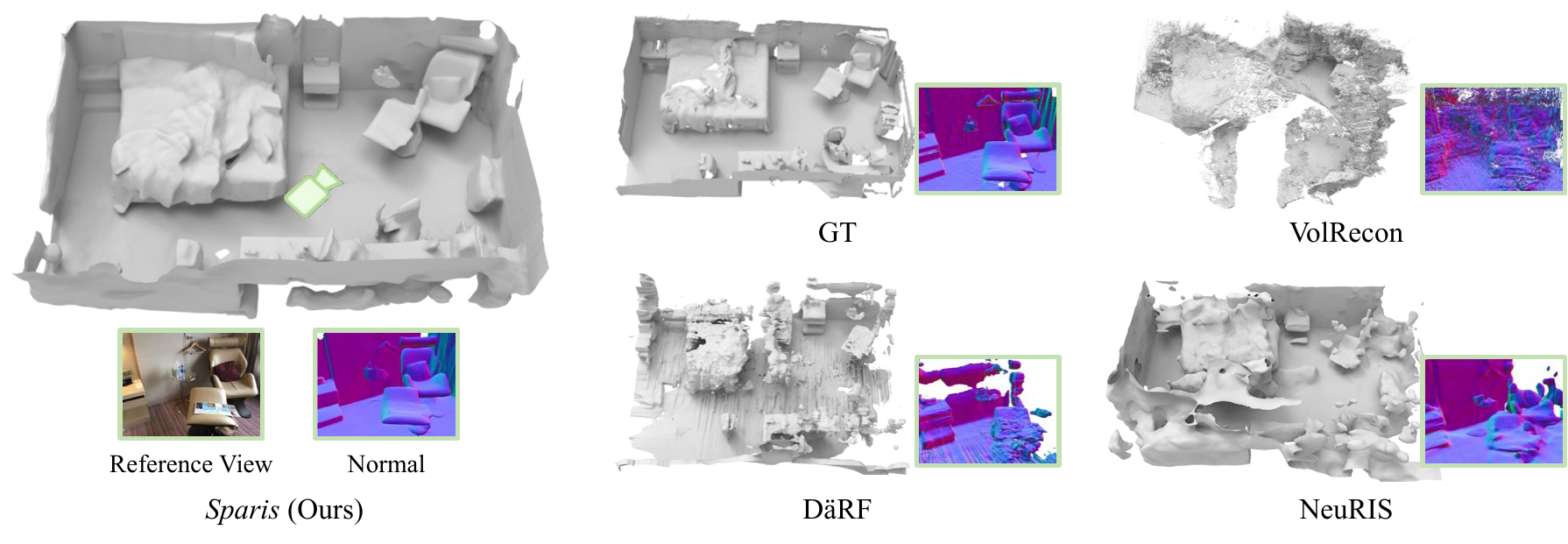}
    \caption{Surface reconstruction results from sparse views of an indoor scene. Our method Sparis outperforms in addressing challenges such as missing reconstruction details (NeuRIS), uneven surface (DäRF), and spatial noise (VolRecon).} % \cite{wang2022neuris} \cite{darf} \cite{volrecon}
    \label{fig:first}
\end{figure*}
\section{Introduction}
\label{sec:intro}
Reconstructing indoor 3D geometry from multi-view images is a significant task in the field of computer vision and graphics. Due to the sparse nature of indoor image acquisition, traditional Multi-View Stereo (MVS) methods \cite{yao2018mvsnet,ding2022transmvsnet} often face challenges in generating satisfactory results when overlap is limited.

Recently, with the emergence of Neural Radiance Fields (NeRF) \cite{nerf} technology, implicit scene representations have injected new vitality into multi-view reconstruction of 3D scenes. Several works \cite{neus, volsdf} utilize Signed Distance Functions (SDF) as a geometric representation and employ a neural rendering pipeline to accurately learn the geometry of scenes from multi-view images. Although they have gained considerable advancement in indoor scene reconstruction, it is still challenged by texture-less regions (e.g., walls, floors, ceilings) and complex object layouts. To solve these issues, subsequent works leverage structural constraints \cite{manhattan, s3precon,infonorm} or general-purpose monocular priors \cite{monosdf, wang2022neuris, liang2023helixsurf} to provide more comprehensive supervision of depth and normal, further enhancing the quality of reconstruction. However, reliable reconstruction results always rely on dense input views. When only sparse views are provided, the performance of these methods significantly decreases.

% 这里应该是说，尽管近年来出现了室内稀疏新视角合成和针对一般物体层面的稀疏表面重建，他们仍然无法解决这一问题
% With the advancement of neural rendering techniques, the novel view synthesis from sparse indoor views has emerged as a pivotal area of research within the neural rendering community. DDP-NeRF \cite{ddpnerf} successfully conducts new view synthesis in indoor scenes with a minimal set of just over ten training images, by employing dense depth priors to effectively mitigate depth noise. SCADE \cite{scade} and DäRF \cite{darf} have refined monocular depth priors, reducing inherent ambiguities to serve as improved constraints for scene reconstruction. However, rendering-oriented methods lack clear geometric representation, thus failing to capture accurate geometry through implicit fields.

% Some recent methods \cite{sparseneus, volrecon, s-volsdf, neusurf, divinet} aim to tackle the challenge of sparse view reconstruction, focusing on object-centric datasets such as DTU \cite{dtu} and BlendedMVS \cite{blendedmvs}. While these methods excel in object-centric sparse reconstructions, they underperform in sparse view indoor reconstruction.

Two categories of approaches have provided inspiration for addressing the problem of indoor sparse-view reconstruction, yet cannot resolve this issue. Indoor sparse novel view synthesis methods \cite{ddpnerf, scade, darf} improve rendering quality by employing dense depth priors or refined monocular priors, but fail to capture accurate geometry for lacking of clear geometric representation. Object level sparse reconstruction methods \cite{sparseneus, volrecon, s-volsdf, neusurf} enhance the feature extraction capabilities of neural fields while underperform in large and complex indoor scenes.

In this work, we adopt SDF for geometric representation, revisiting the paradigms of prior-based neural implicit learning under sparse settings. We notice that enforcing monocular depth supervision diminishes the reconstruction quality due to the inability to calibrate depth scale within sparse views. In addressing this challenge, we leverage matching information between images to obtain more reliable absolute depth prior. Additionally, to further ensure consistency between views, we introduce a reprojection loss, which optimizes the reconstruction geometry surface based on matching relationships. As our matching relationships are entirely determined by the matching network, matching errors may impact the accuracy of our priors. We designed a matching mechanism consisting of a matching angle filter and an epipolar weight function. The matching angle filter calculates the angular score between views and can filter out severe bias introduced by matching errors in nearby perspectives.
The epipolar weight function calculates the Sampson Distance for matched pixels within the corresponding images, and quantitatively assesses their correspondence in 3D space, enhancing the overall accuracy of reconstruction. As shown in Figure \ref{fig:first}, our method can achieve more complete and detailed surface reconstruction, compared with previous approaches. We highlight our key contributions as follows.%all current approaches encounter various geometric challenges that prevent them from completing indoor reconstruction tasks with sparse views.
\begin{itemize}
\item We propose \textit{Sparis}, a novel surface reconstruction method that utilizes correspondence information between images for indoor sparse-view reconstruction. Our method leverages pixel-pair information for depth optimization and reprojection losses to refine the surface.
\item We develop matching optimization strategies aimed at minimizing the effects of matching inaccuracies, ensuring more reliable depth and reprojection alignments.
\item Our extensive evaluations on both real-world and synthetic datasets show that \textit{Sparis} achieves superior performance over current leading indoor reconstruction methods with sparse views.
\end{itemize}

\section{Related Works}
\subsection{Novel View Synthesis for Indoor Scenes}
Synthesizing images from novel viewpoints of a scene within a set of images has long attracted attention in the field of computer vision. Recently, Neural Radiance Fields (NeRF) \cite{nerf} as a neural implicit representation method, achieves high-quality and view-dependent rendering through a volume rendering pipeline. Based on NeRF, many studies have made improvements in rendering speed \cite{kilonerf, plenoctrees, dvgo, instant-ngp, wenyuan}, quality \cite{mipnerf, mipnerf360, next, han2024binocular}, and generalizability \cite{mvsnerf, geonerf, enhancingnerf}. Apart from architectural improvements in universal conditions, some researchers have focused on specific categories of scene reconstruction, such as indoor \cite{parf, surfelnerf}, outdoor \cite{nerflets, neo360}, underwater \cite{seathru}, satellite \cite{satnerf}, and urban environments \cite{blocknerf, urbannerf, meganerf}, aiming to achieve higher rendering quality within these distinct settings. However, in indoor scenes, challenges often arise due to a limited number of images and small coverage areas. DDP-NeRF \cite{ddpnerf} trained a dense depth prior from a large indoor dataset to constrain the small-convergence NeRF optimization. DäRF \cite{darf} and SCADE \cite{scade} improved the ambiguity and scale issues of the SOTA monocular depth model priors, leading to more accurate depth supervision and thereby enhancing the rendering effects. Although significant progress has been made in synthesizing novel views for sparse indoor scenes, these results fail to achieve the reconstruction geometry under sparse views due to the lack of accurate geometric representations, such as Signed Distance Functions (SDF).

\begin{figure*}[!ht]
    \centering
    \includegraphics[width=\textwidth]{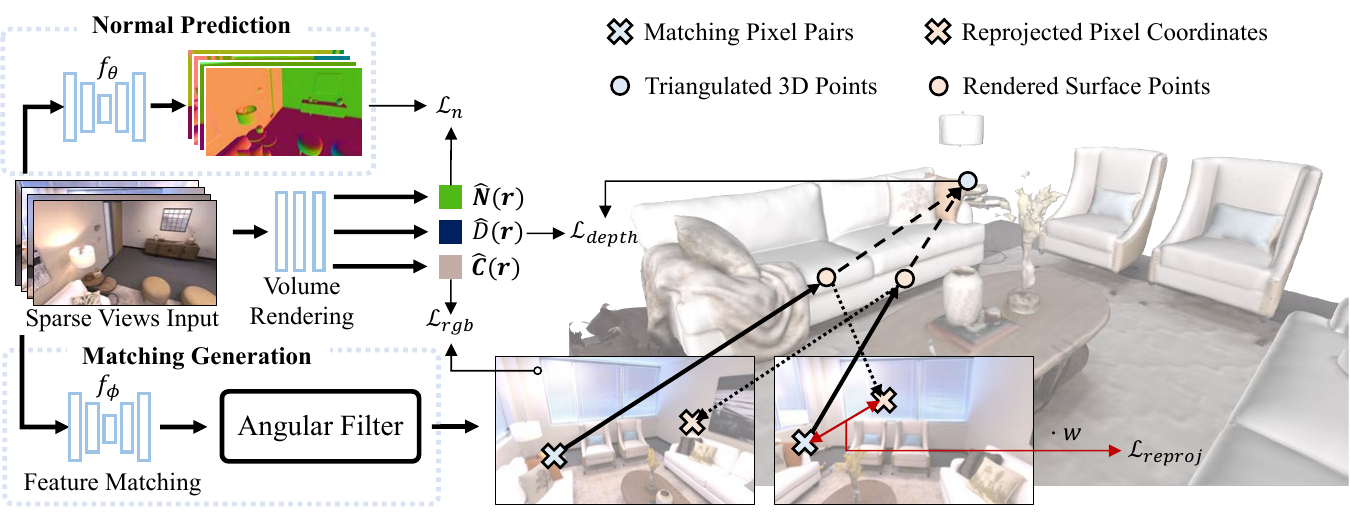}
    \caption{The overview of Sparis. Given sparse indoor images, the reconstruction of 3D surfaces is achieved via a 2-stage process: (1) Pre-processing: estimated normal maps and matching pixel pairs are derived respectively using a pre-trained normal prediction network $f_\theta$ and a feature matching network $f_\phi$; (2) Training with priors: the neural rendering procedure is optimized with inter-image depth priors, cross-view reprojection and monocular normal priors, generating complete and detailed geometry.}
    \label{fig:pipeline}
\end{figure*}
\subsection{Geometry Reconstruction for Indoor Scenes}
Reconstructing geometric surfaces from multiple viewpoints is relatively straightforward for a single object with dense views.
Inspired by NeRF, NeuS \cite{neus} and VolSDF \cite{volsdf}  utilized a volumetric rendering pipeline to learn the neural implicit surfaces of objects from multi-view images. HelixSurf \cite{liang2023helixsurf} and Neus2 \cite{neus2} adopted dense grid feature coding like instant-ngp \cite{instant-ngp} to accelerate the reconstruction process. However, they are generic, object-centric methods that perform poorly in scenes with many untextured areas and significant lighting variations. To tackle the challenges of indoor scene reconstruction, MonoSDF \cite{monosdf} and NeuRIS \cite{wang2022neuris} introduced 2D pre-trained models as priors, effectively dealing with the issues in reconstructing untextured areas. Manhattan-SDF \cite{manhattan} and S$^3$P \cite{s3precon} did not employ geometric priors from 2D images; instead, they drew upon the laws of the physical world to design constraints on surface normals for indoor scenes. However, current indoor reconstruction works still demand a high number of images, often requiring hundreds of images to achieve satisfactory reconstruction results. Recently, some works \cite{sparseneus, volrecon, neusurf, s-volsdf, xu2023c2f2neus} have attempted to perform implicit surface reconstruction with sparse views. Yet, these studies primarily concentrate on object-centric reconstruction with few views, overlooking indoor scenes. With more objects and less view overlap in the scene, reconstructing under sparse views grows increasingly difficult.

\section{Method}
In this study, we aim to reconstruct the fidelity surface $\mathcal{S}$ of an indoor scene from a limited set of images $\mathcal{I} = \{I_i\ |\ i\in1,\dots,M\}$ and camera poses $\mathcal{T} = \{T_i\ |\ i\in1,\dots,M\}$. We introduce \textit{Sparis}, a neural surface reconstruction approach optimized for sparse view inputs, as illustrated in Figure \ref{fig:pipeline}.
\subsection{Neural Implicit Surface Volume Rendering}
We model both geometry and appearance using SDF and color fields, learned by the differentiable rendering pipeline. Defining the surface geometry of the indoor scene as the zero-level set of SDF $\mathcal{S}=\{\bm{x}\in\mathbb{R}^3\ |\ f(\bm{x})=0\}$, we then adopt VolSDF \cite{volsdf} as our baseline. This allows for the transformation of SDF into volumetric density for volume rendering, with both SDF and color parameterized by two MLPs as VolSDF.

Given a pixel from one image, the ray could be denoted as $\{\bm{r}(t_i) = \bm{o}+t\bm{d}\ |\ t>0\}$, where $\bm{o}$ is the camera center and $\bm{d}$ is the direction of the ray. The rendered color is accumulated by volume rendering with $N$ discrete points:
\begin{equation}
    \hat{\bm{C}}(\bm{r}) = \sum_{i=1}^NT_i\alpha_i\bm{c}_i\ ,
\end{equation}
where $T_i$ is the accumulated transmittance, $\alpha_i$ is the opacity values, as denoted by
\begin{equation}
    T_i = \prod_{j=1}^{i-1}(1-\alpha_i)\ ,\quad
    \alpha_i = 1-\exp\left(-\sigma_i\delta_i\right).
\end{equation}
Following VolSDF, we transform SDF values $s$ to density values $\sigma$ using a learnable parameter $\beta$:
\begin{equation}\label{e:laplace}
\sigma(s) = \begin{cases} \frac{1}{2\beta} \exp\left( \frac{s}{\beta} \right) & \text{if } s\leq 0 \\
\frac{1}{\beta}\left( 1-\frac{1}{2}\exp\left ( -\frac{s}{\beta} \right ) \right) & \text{if } s>0
\end{cases}\ .
\end{equation}
Subsequently, we calculate the depth $\hat{D}(\bm{r})$ and normal $\hat{\bm{N}}(\bm{r})$ at the intersection of the surface with the current ray using the following expressions:
\begin{equation}
\hat{D}(\bm{r}) = \sum_{i=1}^N \, T_i \, \alpha_i \, t_i\ ,\quad
\hat{\bm{N}}(\bm{r}) = \sum_{i=1}^N \, T_i \, \alpha_i \, \hat{\bm{n}}_i \ .
\label{eq:volume_render_dn}
\end{equation}

\subsection{Inter-Image Depth Loss}
MonoSDF \cite{monosdf} represents a foundational work in multi-view indoor reconstruction, introducing the Omnidata \cite{ominidata} depth prior that supplies a wealth of geometric information. To enforce the consistency between the render depth $\hat{D}(\bm{r})$ and monocular depth $\bar{D}(\bm{r})$, It employs a loss function that is invariant to scale:
\begin{equation}
\mathcal{L}_{mono\ depth} = \sum_{\bm{r}\in\mathcal{R}}{\left\lVert\left(w\hat{D}(\bm{r})+q\right)-\bar{D}(\bm{r}) \right\rVert}^2.
\label{eq:depth_prior}
\end{equation}
This means that the relative depth from monocular input needs to be scaled to an absolute scale for geometry supervision. Scale $w$ and shift $q$ are solved with the least-squares criterion in the rendering process. 

However, this strategy can lead to severe depth ambiguity problems, ultimately resulting in the collapse of the reconstructed geometry. This arises from sparse views training process, where the small overlap between sparse views results in only a limited amount of rendering depth being correctly scaled. Global scale and shift are miscalculated, ultimately leading to errors in depth supervision scale.
%An intuitive approach involves directly supervising the geometry using absolute depth, yet the accuracy of current monocular absolute depth estimation methods remains inadequate, with these methods also facing challenges in generalization capabilities. COLMAP \cite{colmap} is considered to be used in the relative depth calculation in previous works. Nonetheless, its effectiveness is limited, offering only marginal improvements to surface reconstructions due to its reliance on computing a few points for images with limited coverage. The completion method \cite{ddpnerf} for sparse depth in COLMAP, while offering more comprehensive depth information, lacks precision and is limited to optimization for novel view rendering.

To resolve this issue, we introduce a 2D feature points matching network to compute correspondence information between sparse views, utilizing this inter-image information along with image poses to acquire more accurate depth priors. Given a pair of images captured from different viewpoints of current scene, marked as $\{I_{a},I_{b}\}$, we can directly obtain the matching pixel pairs $(\bm{p}_{a},\bm{p}_{b})$ along with an associated matching uncertainty by employing the feature matching network $f_\phi$:
\begin{equation}\label{pixel_matching}
\left\{(\bm{p}^i_{a},\bm{p}^i_{b},u^i_{a,b})\ |\ i\in1,\dots,H\right\} =f_\phi(I_{a},I_{b})\ .
\end{equation}
Here, $H$ denotes the quantity of matching pixel pairs. Matching uncertainty $u_{a,b}$ is quantified within the range of $[0,1]$, indicating the confidence of the matching results.

By leveraging the camera poses alongside these matching pixel pairs, it becomes feasible to triangulate the estimated world coordinates $\bm{x}$, thus inferring absolute depth priors $\widetilde{D}(\bm{r})$, as demonstrated in Figure \ref{fig:triangulation} (a). Throughout the training phase, for a given reference view $I_{r}$, we systematically sample a batch of rays $\{\bm{r}_{r}^i\}$ and rays of their corresponding matching pixels $\{\bm{r}^i_{s}\}$ from a source view $I_{s}$. Consequently, the inter-image depth loss can be expressed as
\begin{equation}
\label{depth_loss}
\mathcal{L}_{depth}=\sum_{i}\frac{1}{\widetilde{D}(\bm{r}_{r}^i)}(1-u^i_{r,s})\left|\hat{D}(\bm{r}_{r}^i) -\widetilde{D}(\bm{r}_{r}^i)\right|.
\end{equation}

\subsection{Cross-View Reprojection Loss}
%Although inter-image depth loss provides accurate geometric supervision, it loses the inter-image correlation. 
During the optimization process of neural rendering, we only compute the depth loss for the current image, ensuring one-way accuracy of inter-image information. When the depth loss converges well, we can approximately assume that the correspondence in inter-image relationships has been ensured. However, in each iteration of the neural rendering pipeline, only a small number of pixels from one view are selected, making it challenging to synchronize the depths of one-to-one corresponding pixels in inter-image relationships. 
%This implies that the one-to-one correspondence of pixels in inter-image relationships is not ensured in the depth optimization process. 
To tackle this challenge, we introduce reprojection for optimization.

\begin{figure}[!t]
    \centering
    \includegraphics[width=\columnwidth]{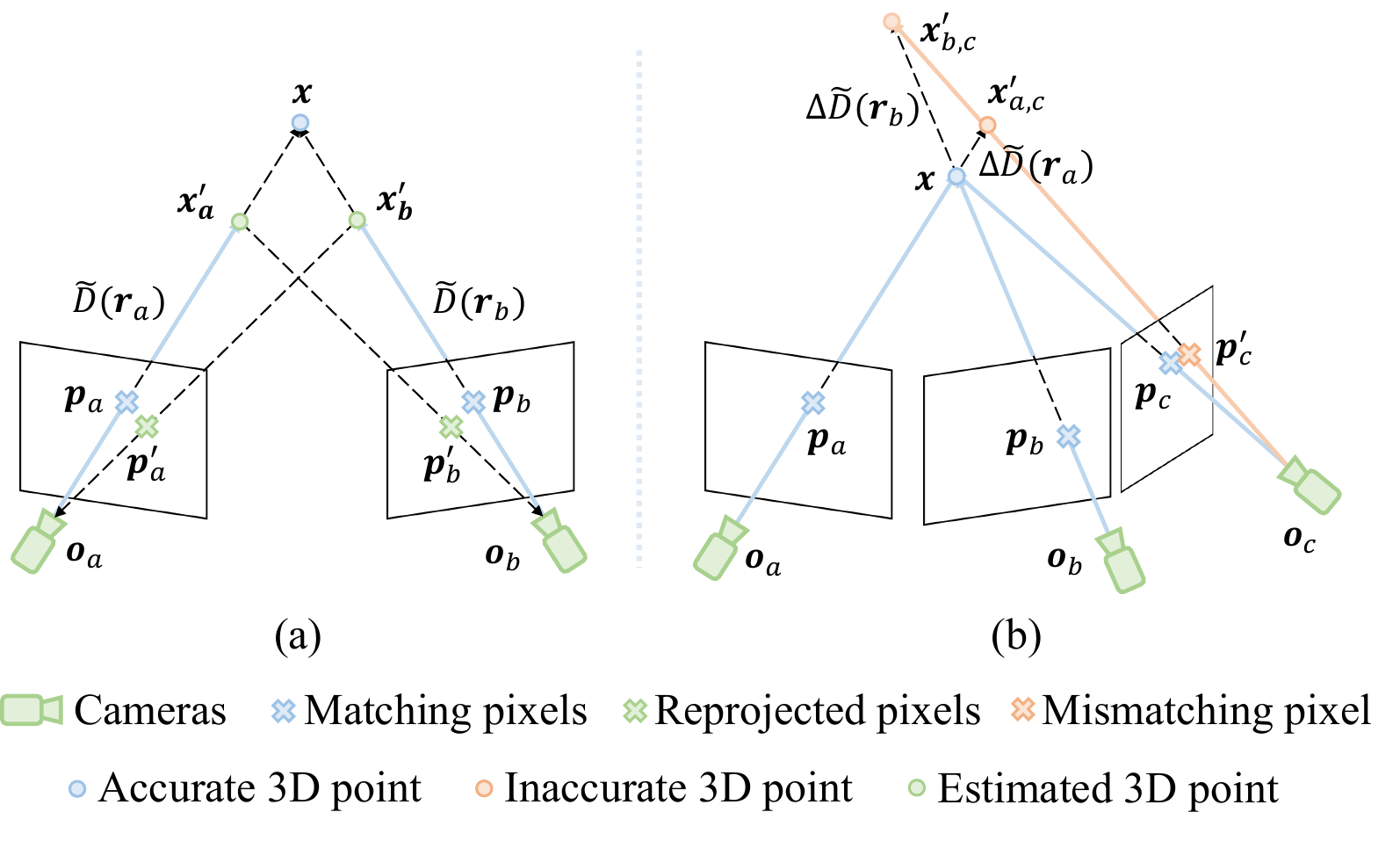}
    \caption{Illustration of matching priors. (a) Using matching pixel pairs, we obtain the triangulated depth $\widetilde{D}$ and reprojected coordinates $\bm{p}^\prime$ from the rendered 3D surface points; (b) Mismatches cause depth estimation errors, especially under minimal translation and angular variations.}
    \label{fig:triangulation}
\end{figure}

As it is shown in Figure \ref{fig:triangulation} (a), considering that the point $\bm{x}^\prime_{a}$ on which a ray intersect with the surface estimated by neural rendering is not coincident with the triangulated point $\bm{x}$, since the error between rendered surfaces and real surfaces exists, an offset is also introduced between the reprojected coordinate $\bm{p}^\prime_{b}$ and $\bm{p}_{b}$ on another view. Given a reference view $I_{r}$ and a source view $I_{s}$, the reprojected coordinate $\bm{p}_{s}^\prime$ from the rendered 3D point of reference view to the source view can be calculated as
\begin{equation}
\bm{p}_{s}^\prime =KP_{s}^{-1}\left(\bm{o}_{r}+\hat{D}(\bm{r}_{r})\cdot\bm{d}(\bm{r}_{r})\right),
\end{equation}
where $K$ denotes camera intrinsic matrix, $P$ represents the camera pose, $\bm{d}$ is the normalized direction of $\bm{r}$, and $\bm{o}$ indicates the world coordinate of the camera viewpoint.

Then, the cross-view reprojection loss is calculated as
\begin{equation}\label{reproj_loss}
\mathcal{L}_{reproj}=\sum_{i}\mathcal(1-u^i_{r,s})\left\Vert\bm{p}^i_{s}-{\bm{p}_{s}^i}^\prime\right\Vert_1,
\end{equation}
$\Vert\cdot\Vert_1$ represents the L1 norm.

\subsection{Matching Optimization Strategies}
%Different from monocular depth estimation, not all pixels would be supervised with depth priors during neural rendering, considering not all pixels in the reference view find a corresponding matching pixel in the source view. In order to fully utilize the inter-image correspondence information, it's significant to select source views that match more. 
Matching networks inherently introduce certain errors, which can lead to geometric inaccuracies and spatial noise. To mitigate these issues, we develop two optimization strategies: angular filter for refining image matching pairs and epipolar weight function for enhancing pixel matching pairs.

\subsubsection{Angular Filter.}
In multi-view geometry, triangulation errors are strongly influenced by the angles between ray pairs. As illustrated in Figure \ref{fig:triangulation} (b), smaller angles result in greater relative errors in depth estimation when mismatches occur. Therefore, relying solely on the number of matching pairs as a metric for source view selection can lead to more inaccurate estimations. To mitigate this, we compute the certainty-weighted average of the normalized direction vectors of the rays at each matching pixel. The score for translation and angular variations of the views is then calculated as
\begin{equation}\label{match_score}
S_{a,b}=1-\cos\left(\sum_{i}(1-u_{a,b}^{i})\bm{d}_{a}^{i}\ ,\ \sum_{i}(1-u_{a,b}^{i}) \bm{d}_{b}^{i}\right),
\end{equation}
where $\bm{d}$ are the normalized direction vectors of rays. For reference view $I_{r}$, the source view is picked as
\begin{equation}\label{source_view}
I_{s} = \arg\max([S_{r,i}-\epsilon>0]\cdot H_{{r,}i})\ ,\ i\neq{r}\ .
\end{equation}
$H$ indicates the number of matching pixel pairs. $[\cdot]$ represents the Iverson bracket.

\begin{figure*}[!ht]
    \centering
    \includegraphics[width=\textwidth]{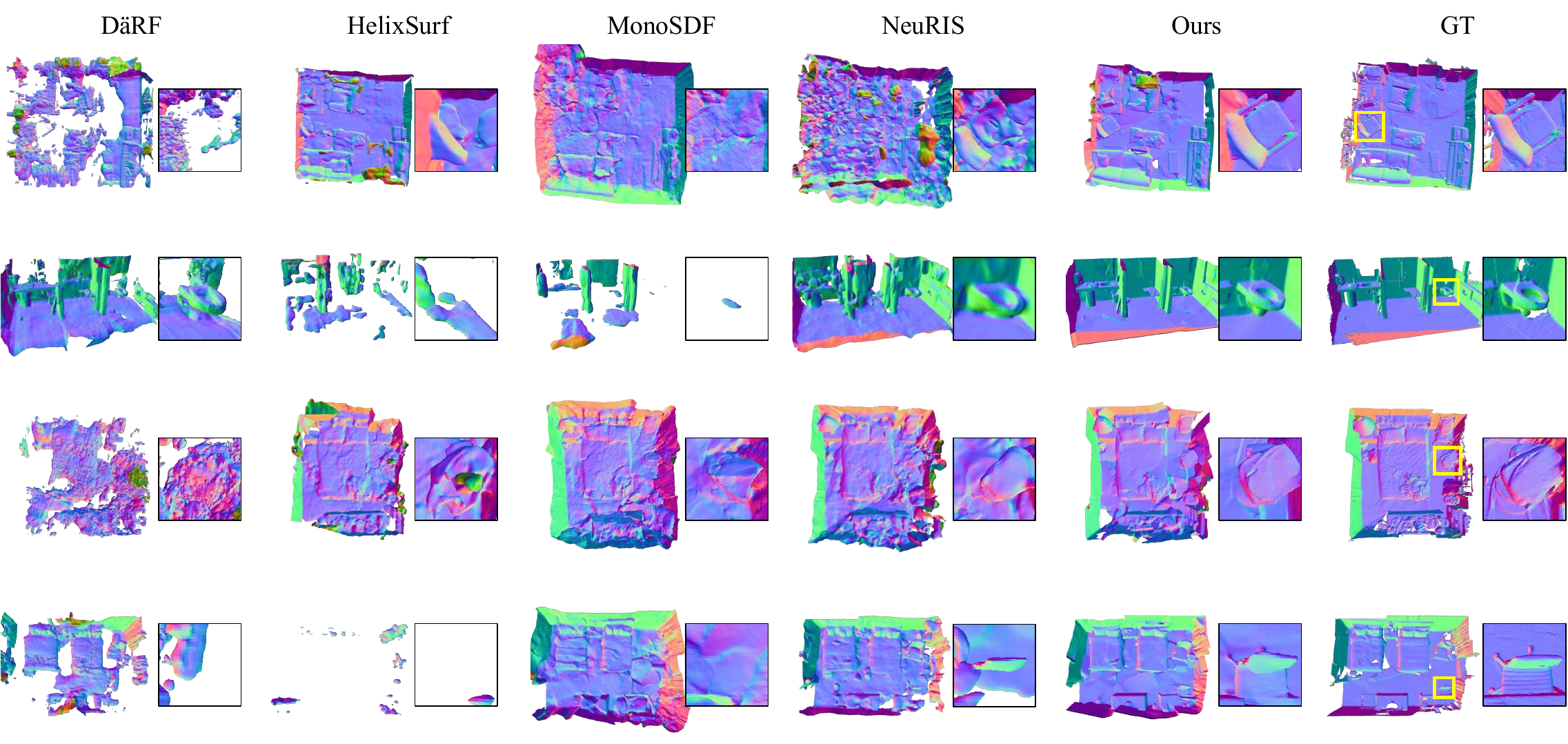}
    \caption{Visual comparisons of 3D reconstruction results on ScanNet with sparse views. The overall top views and the zoom-in views of the marked areas show that our approach produces more complete and fine-grained geometry.}
    \label{fig:comparison}
\end{figure*}
\subsubsection{Epipolar Weight Function.}
Feature matching networks, as data-driven models operating at the image level, often lack verification of multi-view geometric consistency within the scene. Consequently, the matched pixels may not adhere to the correct spatial-geometric relationships, failing to meet the scene's geometric constraints. Ideally, during triangulation, the rays corresponding to a matched pixel pair should intersect at a single point, with the pixels lying on the epipolar lines. To mitigate this limitation, we introduce an epipolar weight, which can be computed as
\begin{equation}\label{sampson_weight}
w_{r,s}^i=\frac{1}{2}\left(1-\mathrm{Sigmoid}\left(\gamma\cdot d_{s}(\bm{p}_{r}^i,\bm{p}_{s}^i)\right)\right).
\end{equation}
$d_{s}$ represents the Sampson Distance, calculated as
\begin{equation}\label{sampson_dist}
d_{s}(\bm{p}_{r},\bm{p}_{s}) =\frac{(\bm{p}_{s}^\top F\bm{p}_{r})^2}{(F\bm{p}_{r})_1^2+(F\bm{p}_{r})_2^2+(F^\top\bm{p}_{s})_1^2+(F^\top\bm{p}_{s})_2^2},
\end{equation}
where $F$ denotes the fundamental matrix between reference view $I_{r}$ and source view $I_{s}$. $(\cdot)_k$ represents the $k$-th element of the vector. Thus, with the consideration of the epipolar weights, $\mathcal{L}_{depth}$ and $\mathcal{L}_{reproj}$ can be rewrote as: 
\begin{equation}\label{new_depth_loss}
\mathcal{L}_{depth}=\sum_{i}\frac{1}{\widetilde{D}(\bm{r}_{r}^i)}(1u^i_{r,s})w_{r,s}^i\left|\hat{D}(\bm{r}_{r}^i)-\widetilde{D}(\bm{r}_{r}^i)\right|,
\end{equation}
\begin{equation}\label{new_reproj_loss}
\mathcal{L}_{reproj}=\sum_{i}\mathcal(1-u^i_{r,s})w_{r,s}^i\left\Vert\bm{p}^i_{s}-\bm{p}_{s}^{i\prime}\right\Vert_1.
\end{equation}

\subsection{Loss Functions}
The overall loss functions are:
\begin{equation}\label{loss_function}
\mathcal{L}=\mathcal{L}_{rgb}+\lambda_{1}\mathcal{L}_{depth}+\lambda_{2}\mathcal{L}_{reproj}+\lambda_{3}\mathcal{L}_{n}+\lambda_{4}\mathcal{L}_{eik},
\end{equation}
where $\mathcal{L}_{depth}$ and $\mathcal{L}_{reproj}$ are the inter-image depth loss and cross-view reprojection loss defined above.

$\mathcal{L}_{rgb}$ is the difference between the rendered and ground-truth pixel colors:
\begin{equation}\label{color_loss}
\mathcal{L}_{rgb}=\frac{1}{|\mathcal{R}|}\sum_{\bm{r}\in{\mathcal{R}}}{\left\Vert\bm{C}(\bm{r})-\hat{\bm{C}}(\bm{r})\right\Vert_1}.
\end{equation}
Similar to NeuRIS \cite{wang2022neuris}, we utilize a pre-trained network $f_\theta$ to predict monocular normals $\bar{\bm{N}}(\bm{r})$, which are then applied to the Normal loss: 
\begin{equation}\label{normal_loss}
\mathcal{L}_{n}=\frac{1}{|\mathcal{R}|}\sum_{\bm{r}\in\mathcal{R}}\left\Vert\hat{\bm{N}}(\bm{r})-\bar{\bm{N}}(\bm{r})\right\Vert_1+\left\Vert 1-\hat{\bm{N}}(\bm{r})^\top\bar{\bm{N}}(\bm{r})\right\Vert_1.
\end{equation}
In line with the previous approaches, we introduce an Eikonal regularization term \cite{IGR} on the random sample points $\mathcal{Y}$ for the SDF field $f(\bm{x})$:
\begin{equation}
\mathcal{L}_{eik}=\frac{1}{|\mathcal{Y}|}\sum_{\bm{x}\in{\mathcal{Y}}}{(\Vert\nabla f(\bm{x})\Vert-1)^2}.
\end{equation}

\section{Experiments and Anaysis}
\subsection{Datasets}
\begin{table*}[!ht]
    \centering
        \begin{tabular}{>{\centering\arraybackslash}p{7cm} >{\centering\arraybackslash}p{1.5cm} >{\centering\arraybackslash}p{1.5cm} >{\centering\arraybackslash}p{1.5cm} >{\centering\arraybackslash}p{1.5cm} >{\centering\arraybackslash}p{1.5cm} >{\centering\arraybackslash}p{1.5cm}}
            \toprule\noalign{\smallskip}
            Method & F-score$\uparrow$ & Acc.$\downarrow$ & Comp.$\downarrow$ & Prec.$\uparrow$ & Recal.$\uparrow$ \\
            % \midrule
            \midrule %\noalign{\smallskip}
            \midrule
            COLMAP \cite{colmap} & 0.161 & \underline{0.179} & 0.583 & 0.284 & 0.124 \\
            TransMVSNet \cite{ding2022transmvsnet} & 0.119 & 0.352 & 0.473 & 0.142 & 0.102 \\
            \cmidrule(l{0.3em}r{0.3em}){1-6}
            DDP-NeRF \cite{ddpnerf} & 0.287 & 0.280 & \underline{0.080} & 0.202 & \underline{0.539} \\
            % SCADE\cite{scade} &  &  & & & \\
            D\"aRF \cite{darf} & 0.295 & 0.273 & 0.127 & 0.242 & 0.393 \\
            \cmidrule(l{0.3em}r{0.3em}){1-6}
            NeuS \cite{neus} & 0.132 & 0.300 & 0.665 & 0.185 & 0.105 \\
            VolRecon \cite{volrecon} & 0.155 & 0.225 & 0.284 & 0.174 & 0.144 \\
            HelixSurf \cite{liang2023helixsurf} & 0.238 & 0.341 & 0.249 & 0.249 & 0.229 \\
            S$^{3}$P \cite{s3precon} & 0.277 & 0.300 & 0.177 & 0.274 & 0.285 \\
            MonoSDF \cite{monosdf} & 0.328 & 0.328 & 0.152 & 0.320 & 0.341 \\
            NeuRIS \cite{wang2022neuris} & \underline{0.464} & 0.180 & 0.082 & \underline{0.445} & 0.488 \\
            \cmidrule(l{0.3em}r{0.3em}){1-6} %\noalign{\smallskip}
            Ours & \textbf{0.647} & \textbf{0.056} & \textbf{0.060} & \textbf{0.666} & \textbf{0.631} \\
            \bottomrule
        \end{tabular}
        \caption{Quantitative comparisons of room-scale surface reconstruction results over 10 scenes of ScanNet with 15-20 input views. The best and the second best results are denoted as bold and underlined, respectively.}
    \label{tab:comparison}
\end{table*}
\begin{table*}[!ht]
    \centering
        \begin{tabular}{>{\centering\arraybackslash}p{5cm} >{\centering\arraybackslash}p{1.5cm} >{\centering\arraybackslash}p{1.5cm} >{\centering\arraybackslash}p{1.5cm} >{\centering\arraybackslash}p{1.5cm} >{\centering\arraybackslash}p{1.5cm} >{\centering\arraybackslash}p{1.5cm}}
            \toprule\noalign{\smallskip}
            Method & F-score$\uparrow$ & Acc.$\downarrow$ & Comp.$\downarrow$ & Prec.$\uparrow$ & Recal.$\uparrow$ \\
            \midrule
            % \midrule %\noalign{\smallskip}
            % VolRecon\cite{volrecon} & & & & & \\
            HelixSurf \cite{liang2023helixsurf} & 0.028 & 0.558 & 0.927 & 0.035 & 0.019 \\
            S$^{3}$P \cite{s3precon} & 0.018 & 0.271 & 2.733 & 0.152 & 0.010 \\
            MonoSDF \cite{monosdf} & \underline{0.454} & 0.081 & \underline{0.139} & \underline{0.497} & \underline{0.423} \\
            NeuRIS \cite{wang2022neuris} & 0.431 & \underline{0.074} & 0.147 & 0.489 & 0.387 \\
            % \cmidrule(l{0.3em}r{0.3em}){1-6} %\noalign{\smallskip}
            Ours & \textbf{0.825} & \textbf{0.031} & \textbf{0.073} & \textbf{0.881} & \textbf{0.777} \\
            \bottomrule
        \end{tabular}
    \caption{Quantitative comparisons of room-scale surface reconstruction results over 8 scenes of Replica with 10 input views. The best and the second best results are denoted as bold and underlined, respectively.}
    \label{tab:replica_comparison}
\end{table*}
\subsubsection{ScanNet.}
ScanNet \cite{dai2017scannet}, a comprehensive real-world dataset, encompasses over 2.5 million views across 1513 scenes, each annotated with 3D camera poses and surface reconstructions.  To evaluate the performance of our algorithm, we adopted the sparse setting used by DDP-NeRF \cite{ddpnerf}, sampling 15 to 20 images per scene at a resolution of $624 \times 468$ for surface reconstruction. 
%For consistency, we used the same three scenes as DDP-NeRF \cite{ddpnerf} and extended our evaluation to an additional seven scenes, thereby broadening the scope of our analysis.

\subsubsection{Replica.}
The Replica dataset \cite{straub2019replica} is notable for its high-quality reconstructions of various indoor environments. To further ascertain the robustness of our approach, we followed the scene selection strategy outlined in \cite{monosdf}, opting for 8 distinct scenes. From each scene, 10 images are uniformly sampled out of 2000, at a resolution of $600 \times 340$ for our experimental dataset. 

\subsection{Implementation Details}
We adopt a similar model architecture as VolSDF \cite{volsdf}. RoMa \cite{roma}, a robust network for dense matching, is adopted as network $f_\phi$ to compute priors between images. We utilize the pre-trained Omnidata \cite{ominidata} as our normal estimation network $f_\theta$ to generate monocular normal priors. All the experiments are conducted on an NVIDIA RTX3090 GPU. More experimental settings and metrics calculations are provided in the supplementary materials.%

%We use Adam optimizer \cite{adam} with learning rate of 5e-4 and train the network with batches of 1024 rays for 200k iterations. $\lambda_1, \lambda_2, \lambda_3, \lambda_4, \gamma, \epsilon$ are set to 0.01, 0.01, 0.05, 0.05, 0.1, 0.001. The optimization can be completed in about 12 hours for each scene. 

\subsection{Comparison}
\begin{table*}[!ht]
    \centering
    \begin{tabular}{>{\centering\arraybackslash}p{1.1cm} >{\centering\arraybackslash}p{1.1cm} >{\centering\arraybackslash}p{1.1cm} >{\centering\arraybackslash}p{1.4cm} >{\centering\arraybackslash}p{1.4cm} >{\centering\arraybackslash}p{1.4cm} >{\centering\arraybackslash}p{1.4cm} >{\centering\arraybackslash}p{1.4cm}}
    \toprule\noalign{\smallskip}
    $\mathcal{L}_{n}$ & $\mathcal{L}_{depth}$ & $\mathcal{L}_{reproj}$ & F-score$\uparrow$ & Acc.$\downarrow $ & Comp.$ \downarrow $  & Prec.$ \uparrow $ &   Recall$\uparrow $ \\
    % \noalign{\smallskip}
    \midrule %\noalign{\smallskip}
    & & & 0.244 & 0.177 & 0.319 & 0.308 & 0.212 \\
    \raisebox{0.6ex}{\scalebox{0.7}{$\sqrt{}$}} & & & 0.253 & 0.375 & 0.225 & 0.250 & 0.258 \\
    \raisebox{0.6ex}{\scalebox{0.7}{$\sqrt{}$}} & \raisebox{0.6ex}{\scalebox{0.7}{$\sqrt{}$}} & & 0.598 & 0.061 & 0.074 & 0.624 & 0.577 \\
    \raisebox{0.6ex}{\scalebox{0.7}{$\sqrt{}$}} & & \raisebox{0.6ex}{\scalebox{0.7}{$\sqrt{}$}} & 0.560 & 0.083 & 0.243 & 0.518 & 0.549 \\
    \raisebox{0.6ex}{\scalebox{0.7}{$\sqrt{}$}} & \raisebox{0.6ex}{\scalebox{0.7}{$\sqrt{}$}} & \raisebox{0.6ex}{\scalebox{0.7}{$\sqrt{}$}} & \textbf{0.647} & \textbf{0.056} & \textbf{0.060} & \textbf{0.666} & \textbf{0.631} \\
    \bottomrule
    \end{tabular}
    \caption{Ablation studies of each component of our method over 10 scenes of ScanNet.}
    \label{tab:ablation_study}
\end{table*}
\subsubsection{ScanNet.}
We compare our approach with various types of indoor reconstruction methods on ScanNet dataset: (1) MVS reconstruction methods: COLMAP \cite{colmap}, TransMVSNet \cite{ding2022transmvsnet}; (2) Novel view synthesis methods for sparse-view indoor scenes: DDP-NeRF \cite{ddpnerf}, D\"aRF \cite{darf}; (3) Neural implicit surface methods for sparse-view reconstruction: VolRecon \cite{volrecon}; (4) Neural implicit surface methods for indoor scenes: NeuS \cite{neus}, NeuRIS \cite{wang2022neuris}, MonoSDF \cite{monosdf}, HelixSurf \cite{liang2023helixsurf}, S$^3$P \cite{s3precon}.
\begin{figure}[!ht]
    \centering
    \includegraphics[width=\columnwidth]{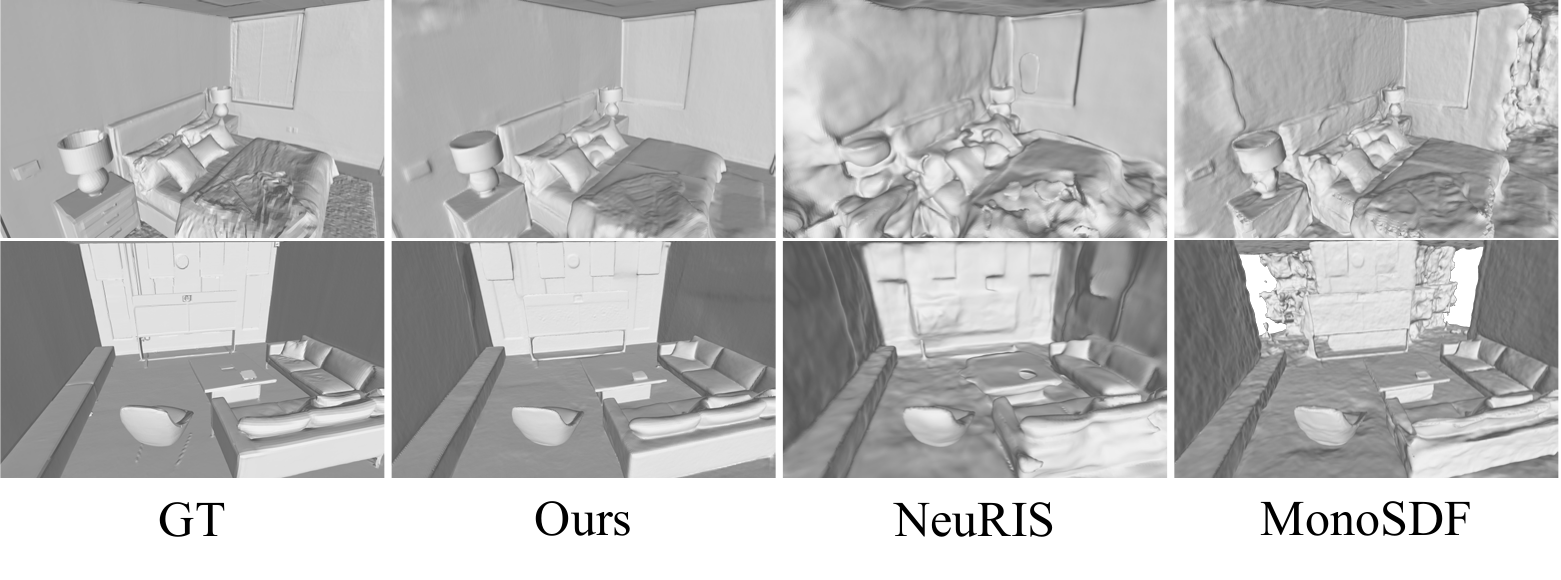}%0.9\textwidth
    \caption{Visual comparisons of 3D reconstruction results on Replica with sparse views.}
    \label{fig:replica}
\end{figure}

To ensure a fair comparison, we fine-tune the experimental setups for specific baselines to maximize their performance. For COLMAP and TransMVSNet, we employ Poisson Reconstruction \cite{poisson} to generate surface meshes from the densely matched point cloud outputs. In the cases of DDP-NeRF and D\"aRF, we utilize the Marching Cube algorithm to create meshes from the learned density fields, applying an appropriately adjusted threshold for optimal results. MonoSDF, under its default hyper-parameter configuration, was unable to produce valid meshes; thus, we modified the weight of the monocular depth loss to 0.001 (originally 0.1) for a more equitable comparison. The quantitative outcomes of this assessment are presented in Table \ref{tab:comparison}. Notably, NeuS was unable to generate valid meshes for 4 scenes, and HelixSurf for 1; hence, we specifically report results for the successfully reconstructed scenes for these methods. Our methodology surpasses all benchmarks, demonstrating a significant improvement. Concurrently, as depicted in Figure \ref{fig:comparison}, previous methods could only generate fragmented and noisy surfaces. In contrast, our technique delivers more visually complete geometries, characterized by smoother surfaces and more refined details.

\subsubsection{Replica.}
As a complementary experiment to validate the robustness on different datasets, we compare our approach with MonoSDF \cite{monosdf}, NeuRIS \cite{wang2022neuris}, HelixSurf \cite{liang2023helixsurf} and S$^3$P \cite{s3precon}. The quantitative comparisons are listed in Table \ref{tab:replica_comparison}, while the visual comparisons are shown in Figure \ref{fig:replica}. The most effective indoor reconstruction methods, MonoSDF and NeuRIS, produce uneven surfaces due to the lack of accurate depth guidance. Our approach exhibits a more pronounced advantage on Replica dataset, characterized by minimal occlusion and precise poses, clearly surpassing several neural indoor surface reconstruction methods.

\subsection{Ablation Study}
To evaluate the effectiveness of the components of our proposed priors, we conduct ablation studies on 5 different settings: (1) Naive neural rendering framework without any introduced prior; (2) Neural rendering framework with normal priors; (3) Ours without cross-view reprojection loss; (4) Ours without inter-image depth loss; (4) Ours: neural rendering framework with normal priors, inter-image depth loss and cross-view reprojection loss.

Table \ref{tab:ablation_study} demonstrates that the monocular normal, as a commonly used form of supervision information for indoor reconstruction, can also improve the reconstruction quality with sparse view inputs. Our inter-image depth loss provides accurate geometric constraints, significantly enhancing the reconstruction quality of fine local details. Furthermore, by ensuring a one-to-one correspondence of matching pixels, the cross-view reprojection loss offers a relaxed yet stable form of supervision. This guarantees inter-view consistency, reduces overfitting in sparse view reconstruction, and ultimately enhances reconstruction quality. The observation readily suggests that the simultaneous application of both constraints not only enhances the geometric quality in each view but also mitigates overfitting in scenarios with few views, leading to a markedly significant improvement compared to the baseline.
\begin{table}[h]
\centering
\small
\begin{tabular}{ccc}%{>{\centering\arraybackslash}p{2.3cm} >{\centering\arraybackslash}p{2cm} >{\centering\arraybackslash}p{1.4cm}}
    \toprule\noalign{\smallskip}
    Epipolar weight & Angular filter & F-score$\uparrow$ \\
    \midrule
    \scalebox{0.85}[1]{$\times$} & \raisebox{0.6ex}{\scalebox{0.7}{$\sqrt{}$}} & 0.617 \\
    \raisebox{0.6ex}{\scalebox{0.7}{$\sqrt{}$}} & \scalebox{0.85}[1]{$\times$} & 0.624 \\
    \raisebox{0.6ex}{\scalebox{0.7}{$\sqrt{}$}} & \raisebox{0.6ex}{\scalebox{0.7}{$\sqrt{}$}} & \textbf{0.647} \\
    \bottomrule    
\end{tabular}
\caption{Quantitative results of ablation study on epipolar weight and angular filter.}
\label{Epipolar ablation}
\end{table}

\begin{figure}[!t]
    \centering
    \includegraphics[width=\columnwidth]{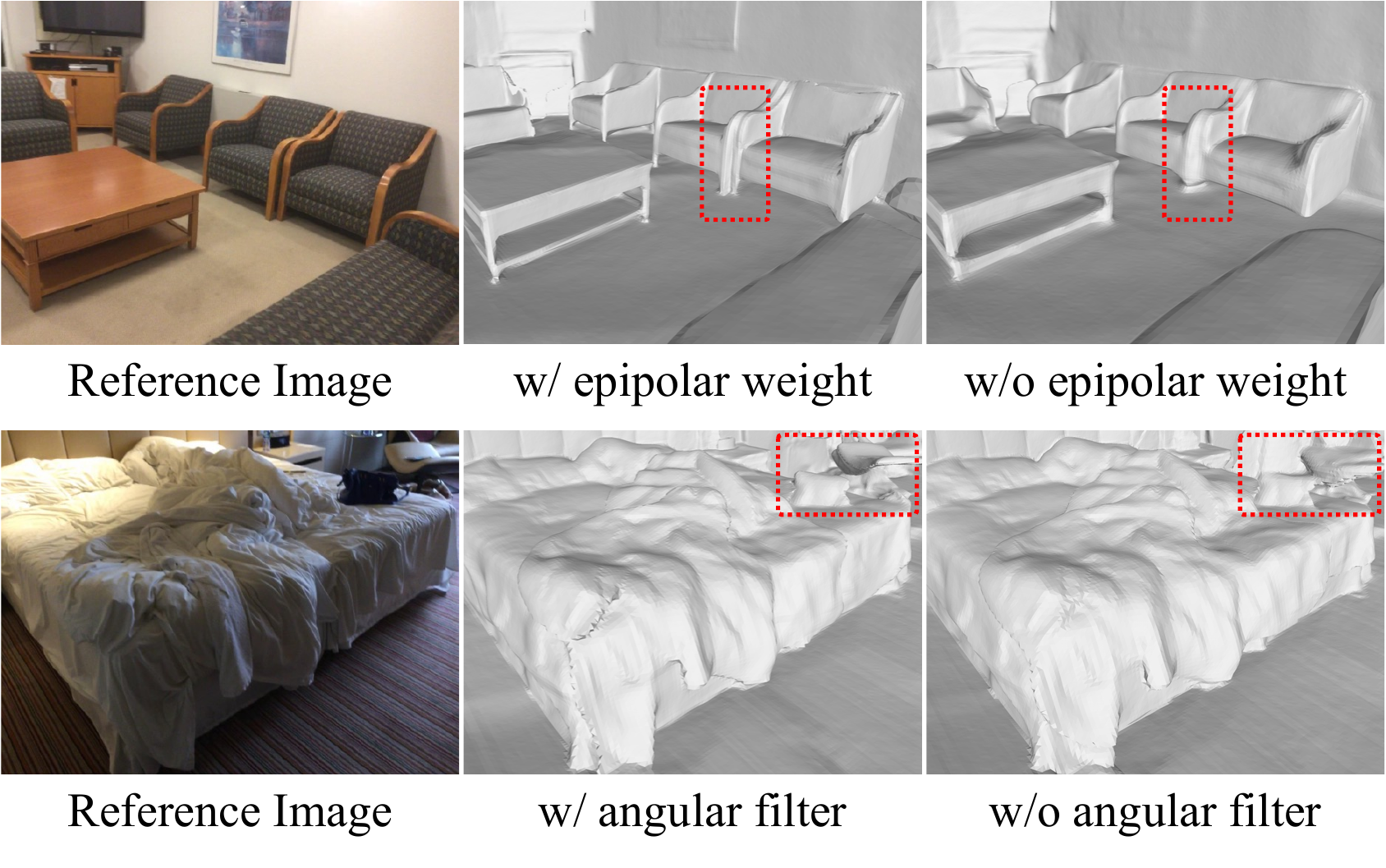}
    \caption{Our reconstruction results with or without matching optimization strategies.}
    \label{fig:epipolar}
\end{figure}

To validate the effectiveness of the matching optimization strategies we propose, we conducted ablation studies on epipolar weight function and angular filter, respectively. As shown in Table \ref{Epipolar ablation}% and Table \ref{Angular filter ablation}
, both strategies significantly enhance geometric reconstruction. This demonstrates that our method is resilient to noise in the matching network. To intuitively demonstrate the effectiveness of our strategies, we visualize the ablation experiments for a scene from ScanNet, as shown in Figure \ref{fig:epipolar}. Owing to the similar textures of the chair legs, the matching network features exhibit high similarity. Without the incorporation of the epipolar weight, the geometric structure of the chairs in the reconstruction degrades, leading to difficulties in distinguishing between different chairs. And it is evident that angular filter is capable of improving the quality of reconstruction at local details by eliminating view pairs with significant error influences.

\section{Conclusion}
We introduce a novel neural implicit surface reconstruction approach for 3D indoor scenes from sparse views. Our method exploits inter-image matching information and utilizes triangulation to provide more accurate depth information than monocular depth, thereby enhancing the stability of the reconstruction process. In addition, we design a projection loss based on pixel-to-pixel matching relationships in the images to ensure consistency across views. To refine accuracy further, we design an angular filter and an epipolar weight function. This helps remove wrong potential matches that might harm the final results. Extensive experiments demonstrate that our method outperforms all existing indoor reconstruction approaches. With only a limited number of views available, we achieve satisfactory reconstruction results on both real and synthetic datasets. 

\section{Acknowledgments}
The corresponding author is Ge Gao. This work was supported by Beijing Science and Technology Program (Z231100001723014).

\bibliography{aaai25}

\begin{thebibliography}{48}
\providecommand{\natexlab}[1]{#1}

\bibitem[{Barron et~al.(2021)Barron, Mildenhall, Tancik, Hedman, Martin-Brualla, and Srinivasan}]{mipnerf}
Barron, J.~T.; Mildenhall, B.; Tancik, M.; Hedman, P.; Martin-Brualla, R.; and Srinivasan, P.~P. 2021.
\newblock Mip-nerf: A multiscale representation for anti-aliasing neural radiance fields.
\newblock In \emph{Proceedings of the IEEE/CVF International Conference on Computer Vision}, 5855--5864.

\bibitem[{Barron et~al.(2022)Barron, Mildenhall, Verbin, Srinivasan, and Hedman}]{mipnerf360}
Barron, J.~T.; Mildenhall, B.; Verbin, D.; Srinivasan, P.~P.; and Hedman, P. 2022.
\newblock Mip-nerf 360: Unbounded anti-aliased neural radiance fields.
\newblock In \emph{Proceedings of the IEEE/CVF Conference on Computer Vision and Pattern Recognition}, 5470--5479.

\bibitem[{Chen et~al.(2021)Chen, Xu, Zhao, Zhang, Xiang, Yu, and Su}]{mvsnerf}
Chen, A.; Xu, Z.; Zhao, F.; Zhang, X.; Xiang, F.; Yu, J.; and Su, H. 2021.
\newblock Mvsnerf: Fast generalizable radiance field reconstruction from multi-view stereo.
\newblock In \emph{Proceedings of the IEEE/CVF International Conference on Computer Vision}, 14124--14133.

\bibitem[{Cong et~al.(2023)Cong, Liang, Wang, Fan, Chen, Varma, Wang, and Wang}]{enhancingnerf}
Cong, W.; Liang, H.; Wang, P.; Fan, Z.; Chen, T.; Varma, M.; Wang, Y.; and Wang, Z. 2023.
\newblock Enhancing nerf akin to enhancing llms: Generalizable nerf transformer with mixture-of-view-experts.
\newblock In \emph{Proceedings of the IEEE/CVF International Conference on Computer Vision}, 3193--3204.

\bibitem[{Dai et~al.(2017)Dai, Chang, Savva, Halber, Funkhouser, and Nie{\ss}ner}]{dai2017scannet}
Dai, A.; Chang, A.~X.; Savva, M.; Halber, M.; Funkhouser, T.; and Nie{\ss}ner, M. 2017.
\newblock Scannet: Richly-annotated 3d reconstructions of indoor scenes.
\newblock In \emph{Proceedings of the IEEE conference on computer vision and pattern recognition}, 5828--5839.

\bibitem[{Ding et~al.(2022)Ding, Yuan, Zhu, Zhang, Liu, Wang, and Liu}]{ding2022transmvsnet}
Ding, Y.; Yuan, W.; Zhu, Q.; Zhang, H.; Liu, X.; Wang, Y.; and Liu, X. 2022.
\newblock Transmvsnet: Global context-aware multi-view stereo network with transformers.
\newblock In \emph{Proceedings of the IEEE/CVF conference on computer vision and pattern recognition}, 8585--8594.

\bibitem[{Edstedt et~al.(2023)Edstedt, Sun, B{\"o}kman, Wadenb{\"a}ck, and Felsberg}]{roma}
Edstedt, J.; Sun, Q.; B{\"o}kman, G.; Wadenb{\"a}ck, M.; and Felsberg, M. 2023.
\newblock RoMa: Revisiting Robust Losses for Dense Feature Matching.
\newblock \emph{arXiv preprint arXiv:2305.15404}.

\bibitem[{Eftekhar et~al.(2021)Eftekhar, Sax, Malik, and Zamir}]{ominidata}
Eftekhar, A.; Sax, A.; Malik, J.; and Zamir, A. 2021.
\newblock Omnidata: A Scalable Pipeline for Making Multi-Task Mid-Level Vision Datasets From 3D Scans.
\newblock In \emph{Proceedings of the IEEE/CVF International Conference on Computer Vision (ICCV)}, 10786--10796.

\bibitem[{Gao, Cao, and Shan(2023)}]{surfelnerf}
Gao, Y.; Cao, Y.-P.; and Shan, Y. 2023.
\newblock SurfelNeRF: Neural Surfel Radiance Fields for Online Photorealistic Reconstruction of Indoor Scenes.
\newblock In \emph{Proceedings of the IEEE/CVF Conference on Computer Vision and Pattern Recognition (CVPR)}, 108--118.

\bibitem[{Gropp et~al.(2020)Gropp, Yariv, Haim, Atzmon, and Lipman}]{IGR}
Gropp, A.; Yariv, L.; Haim, N.; Atzmon, M.; and Lipman, Y. 2020.
\newblock Implicit geometric regularization for learning shapes.
\newblock \emph{arXiv preprint arXiv:2002.10099}.

\bibitem[{Guo et~al.(2022)Guo, Peng, Lin, Wang, Zhang, Bao, and Zhou}]{manhattan}
Guo, H.; Peng, S.; Lin, H.; Wang, Q.; Zhang, G.; Bao, H.; and Zhou, X. 2022.
\newblock Neural 3d scene reconstruction with the manhattan-world assumption.
\newblock In \emph{Proceedings of the IEEE/CVF Conference on Computer Vision and Pattern Recognition}, 5511--5520.

\bibitem[{Han et~al.(2024)Han, Zhou, Liu, and Han}]{han2024binocular}
Han, L.; Zhou, J.; Liu, Y.-S.; and Han, Z. 2024.
\newblock Binocular-Guided 3D Gaussian Splatting with View Consistency for Sparse View Synthesis.
\newblock In \emph{Advances in Neural Information Processing Systems (NeurIPS)}.

\bibitem[{Huang et~al.(2023)Huang, Wu, Zhou, Gao, Gu, and Liu}]{neusurf}
Huang, H.; Wu, Y.; Zhou, J.; Gao, G.; Gu, M.; and Liu, Y. 2023.
\newblock NeuSurf: On-Surface Priors for Neural Surface Reconstruction from Sparse Input Views.
\newblock \emph{arXiv preprint arXiv:2312.13977}.

\bibitem[{Irshad et~al.(2023)Irshad, Zakharov, Liu, Guizilini, Kollar, Gaidon, Kira, and Ambrus}]{neo360}
Irshad, M.~Z.; Zakharov, S.; Liu, K.; Guizilini, V.; Kollar, T.; Gaidon, A.; Kira, Z.; and Ambrus, R. 2023.
\newblock Neo 360: Neural fields for sparse view synthesis of outdoor scenes.
\newblock In \emph{Proceedings of the IEEE/CVF International Conference on Computer Vision}, 9187--9198.

\bibitem[{Johari, Lepoittevin, and Fleuret(2022)}]{geonerf}
Johari, M.~M.; Lepoittevin, Y.; and Fleuret, F. 2022.
\newblock Geonerf: Generalizing nerf with geometry priors.
\newblock In \emph{Proceedings of the IEEE/CVF Conference on Computer Vision and Pattern Recognition}, 18365--18375.

\bibitem[{Kazhdan, Bolitho, and Hoppe(2006)}]{poisson}
Kazhdan, M.; Bolitho, M.; and Hoppe, H. 2006.
\newblock Poisson surface reconstruction.
\newblock In \emph{Proceedings of the fourth Eurographics symposium on Geometry processing}, volume~7, 0.

\bibitem[{Levy et~al.(2023)Levy, Peleg, Pearl, Rosenbaum, Akkaynak, Korman, and Treibitz}]{seathru}
Levy, D.; Peleg, A.; Pearl, N.; Rosenbaum, D.; Akkaynak, D.; Korman, S.; and Treibitz, T. 2023.
\newblock SeaThru-NeRF: Neural Radiance Fields in Scattering Media.
\newblock In \emph{Proceedings of the IEEE/CVF Conference on Computer Vision and Pattern Recognition}, 56--65.

\bibitem[{Liang et~al.(2023)Liang, Huang, Ding, and Jia}]{liang2023helixsurf}
Liang, Z.; Huang, Z.; Ding, C.; and Jia, K. 2023.
\newblock HelixSurf: A Robust and Efficient Neural Implicit Surface Learning of Indoor Scenes with Iterative Intertwined Regularization.
\newblock In \emph{Proceedings of the IEEE/CVF Conference on Computer Vision and Pattern Recognition}, 13165--13174.

\bibitem[{Long et~al.(2022)Long, Lin, Wang, Komura, and Wang}]{sparseneus}
Long, X.; Lin, C.; Wang, P.; Komura, T.; and Wang, W. 2022.
\newblock Sparseneus: Fast generalizable neural surface reconstruction from sparse views.
\newblock In \emph{European Conference on Computer Vision}, 210--227. Springer.

\bibitem[{Mar{\'\i}, Facciolo, and Ehret(2022)}]{satnerf}
Mar{\'\i}, R.; Facciolo, G.; and Ehret, T. 2022.
\newblock Sat-nerf: Learning multi-view satellite photogrammetry with transient objects and shadow modeling using rpc cameras.
\newblock In \emph{Proceedings of the IEEE/CVF Conference on Computer Vision and Pattern Recognition}, 1311--1321.

\bibitem[{Mildenhall et~al.(2020)Mildenhall, Srinivasan, Tancik, Barron, Ramamoorthi, and Ng}]{nerf}
Mildenhall, B.; Srinivasan, P.~P.; Tancik, M.; Barron, J.~T.; Ramamoorthi, R.; and Ng, R. 2020.
\newblock NeRF: Representing Scenes as Neural Radiance Fields for View Synthesis.
\newblock In Vedaldi, A.; Bischof, H.; Brox, T.; and Frahm, J.-M., eds., \emph{Computer Vision -- ECCV 2020}, 405--421. Cham: Springer International Publishing.
\newblock ISBN 978-3-030-58452-8.

\bibitem[{M\"{u}ller et~al.(2022)M\"{u}ller, Evans, Schied, and Keller}]{instant-ngp}
M\"{u}ller, T.; Evans, A.; Schied, C.; and Keller, A. 2022.
\newblock Instant neural graphics primitives with a multiresolution hash encoding.
\newblock \emph{ACM Trans. Graph.}, 41(4).

\bibitem[{Reiser et~al.(2021)Reiser, Peng, Liao, and Geiger}]{kilonerf}
Reiser, C.; Peng, S.; Liao, Y.; and Geiger, A. 2021.
\newblock Kilonerf: Speeding up neural radiance fields with thousands of tiny mlps.
\newblock In \emph{Proceedings of the IEEE/CVF International Conference on Computer Vision}, 14335--14345.

\bibitem[{Rematas et~al.(2022)Rematas, Liu, Srinivasan, Barron, Tagliasacchi, Funkhouser, and Ferrari}]{urbannerf}
Rematas, K.; Liu, A.; Srinivasan, P.~P.; Barron, J.~T.; Tagliasacchi, A.; Funkhouser, T.; and Ferrari, V. 2022.
\newblock Urban radiance fields.
\newblock In \emph{Proceedings of the IEEE/CVF Conference on Computer Vision and Pattern Recognition}, 12932--12942.

\bibitem[{Ren et~al.(2023)Ren, Zhang, Pollefeys, S{\"u}sstrunk, and Wang}]{volrecon}
Ren, Y.; Zhang, T.; Pollefeys, M.; S{\"u}sstrunk, S.; and Wang, F. 2023.
\newblock Volrecon: Volume rendering of signed ray distance functions for generalizable multi-view reconstruction.
\newblock In \emph{Proceedings of the IEEE/CVF Conference on Computer Vision and Pattern Recognition}, 16685--16695.

\bibitem[{Roessle et~al.(2022)Roessle, Barron, Mildenhall, Srinivasan, and Nie{\ss}ner}]{ddpnerf}
Roessle, B.; Barron, J.~T.; Mildenhall, B.; Srinivasan, P.~P.; and Nie{\ss}ner, M. 2022.
\newblock Dense depth priors for neural radiance fields from sparse input views.
\newblock In \emph{Proceedings of the IEEE/CVF Conference on Computer Vision and Pattern Recognition}, 12892--12901.

\bibitem[{Schonberger and Frahm(2016)}]{colmap}
Schonberger, J.~L.; and Frahm, J.-M. 2016.
\newblock Structure-from-motion revisited.
\newblock In \emph{Proceedings of the IEEE conference on computer vision and pattern recognition}, 4104--4113.

\bibitem[{Song et~al.(2023)Song, Park, An, Cho, Kwak, Cho, and Kim}]{darf}
Song, J.; Park, S.; An, H.; Cho, S.; Kwak, M.-S.; Cho, S.; and Kim, S. 2023.
\newblock DäRF: Boosting Radiance Fields from Sparse Inputs with Monocular Depth Adaptation.
\newblock arXiv:2305.19201.

\bibitem[{Straub et~al.(2019)Straub, Whelan, Ma, Chen, Wijmans, Green, Engel, Mur-Artal, Ren, Verma et~al.}]{straub2019replica}
Straub, J.; Whelan, T.; Ma, L.; Chen, Y.; Wijmans, E.; Green, S.; Engel, J.~J.; Mur-Artal, R.; Ren, C.; Verma, S.; et~al. 2019.
\newblock The Replica dataset: A digital replica of indoor spaces.
\newblock \emph{arXiv preprint arXiv:1906.05797}.

\bibitem[{Sun, Sun, and Chen(2022)}]{dvgo}
Sun, C.; Sun, M.; and Chen, H.-T. 2022.
\newblock Direct voxel grid optimization: Super-fast convergence for radiance fields reconstruction.
\newblock In \emph{Proceedings of the IEEE/CVF Conference on Computer Vision and Pattern Recognition}, 5459--5469.

\bibitem[{Tancik et~al.(2022)Tancik, Casser, Yan, Pradhan, Mildenhall, Srinivasan, Barron, and Kretzschmar}]{blocknerf}
Tancik, M.; Casser, V.; Yan, X.; Pradhan, S.; Mildenhall, B.; Srinivasan, P.~P.; Barron, J.~T.; and Kretzschmar, H. 2022.
\newblock Block-nerf: Scalable large scene neural view synthesis.
\newblock In \emph{Proceedings of the IEEE/CVF Conference on Computer Vision and Pattern Recognition}, 8248--8258.

\bibitem[{Turki, Ramanan, and Satyanarayanan(2022)}]{meganerf}
Turki, H.; Ramanan, D.; and Satyanarayanan, M. 2022.
\newblock Mega-nerf: Scalable construction of large-scale nerfs for virtual fly-throughs.
\newblock In \emph{Proceedings of the IEEE/CVF Conference on Computer Vision and Pattern Recognition}, 12922--12931.

\bibitem[{Uy et~al.(2023)Uy, Martin-Brualla, Guibas, and Li}]{scade}
Uy, M.~A.; Martin-Brualla, R.; Guibas, L.; and Li, K. 2023.
\newblock SCADE: NeRFs from Space Carving with Ambiguity-Aware Depth Estimates.
\newblock In \emph{Proceedings of the IEEE/CVF Conference on Computer Vision and Pattern Recognition}, 16518--16527.

\bibitem[{Wang et~al.(2022{\natexlab{a}})Wang, Wang, Long, Theobalt, Komura, Liu, and Wang}]{wang2022neuris}
Wang, J.; Wang, P.; Long, X.; Theobalt, C.; Komura, T.; Liu, L.; and Wang, W. 2022{\natexlab{a}}.
\newblock Neuris: Neural reconstruction of indoor scenes using normal priors.
\newblock In \emph{European Conference on Computer Vision}, 139--155. Springer.

\bibitem[{Wang et~al.(2021)Wang, Liu, Liu, Theobalt, Komura, and Wang}]{neus}
Wang, P.; Liu, L.; Liu, Y.; Theobalt, C.; Komura, T.; and Wang, W. 2021.
\newblock Neus: Learning neural implicit surfaces by volume rendering for multi-view reconstruction.
\newblock \emph{arXiv preprint arXiv:2106.10689}.

\bibitem[{Wang et~al.(2024)Wang, Dong, Zheng, and Yang}]{infonorm}
Wang, X.; Dong, S.; Zheng, Y.; and Yang, Y. 2024.
\newblock InfoNorm: Mutual Information Shaping of Normals for Sparse-View Reconstruction.
\newblock \emph{arXiv preprint arXiv:2407.12661}.

\bibitem[{Wang et~al.(2023)Wang, Han, Habermann, Daniilidis, Theobalt, and Liu}]{neus2}
Wang, Y.; Han, Q.; Habermann, M.; Daniilidis, K.; Theobalt, C.; and Liu, L. 2023.
\newblock Neus2: Fast learning of neural implicit surfaces for multi-view reconstruction.
\newblock In \emph{Proceedings of the IEEE/CVF International Conference on Computer Vision}, 3295--3306.

\bibitem[{Wang et~al.(2022{\natexlab{b}})Wang, Li, Liu, Dai, and Xia}]{next}
Wang, Y.; Li, Y.; Liu, P.; Dai, T.; and Xia, S.-T. 2022{\natexlab{b}}.
\newblock NeXT: Towards High Quality Neural Radiance Fields via Multi-skip Transformer.
\newblock In \emph{European Conference on Computer Vision}, 69--86. Springer.

\bibitem[{Wu, Graikos, and Samaras(2023)}]{s-volsdf}
Wu, H.; Graikos, A.; and Samaras, D. 2023.
\newblock S-VolSDF: Sparse Multi-View Stereo Regularization of Neural Implicit Surfaces.
\newblock \emph{arXiv preprint arXiv:2303.17712}.

\bibitem[{Xu et~al.(2023)Xu, Guan, Wang, Liu, Zeng, Wang, and Yang}]{xu2023c2f2neus}
Xu, L.; Guan, T.; Wang, Y.; Liu, W.; Zeng, Z.; Wang, J.; and Yang, W. 2023.
\newblock C2F2NeUS: Cascade Cost Frustum Fusion for High Fidelity and Generalizable Neural Surface Reconstruction.
\newblock \emph{arXiv preprint arXiv:2306.10003}.

\bibitem[{Yao et~al.(2018)Yao, Luo, Li, Fang, and Quan}]{yao2018mvsnet}
Yao, Y.; Luo, Z.; Li, S.; Fang, T.; and Quan, L. 2018.
\newblock Mvsnet: Depth inference for unstructured multi-view stereo.
\newblock In \emph{Proceedings of the European conference on computer vision (ECCV)}, 767--783.

\bibitem[{Yariv et~al.(2021)Yariv, Gu, Kasten, and Lipman}]{volsdf}
Yariv, L.; Gu, J.; Kasten, Y.; and Lipman, Y. 2021.
\newblock Volume rendering of neural implicit surfaces.
\newblock \emph{Advances in Neural Information Processing Systems}, 34: 4805--4815.

\bibitem[{Ye et~al.(2023)Ye, Liu, Li, and Yang}]{s3precon}
Ye, B.; Liu, S.; Li, X.; and Yang, M.-H. 2023.
\newblock Self-Supervised Super-Plane for Neural 3D Reconstruction.
\newblock In \emph{Proceedings of the IEEE/CVF Conference on Computer Vision and Pattern Recognition}, 21415--21424.

\bibitem[{Ying et~al.(2023)Ying, Jiang, Zhang, Xu, Yu, Dai, and Fang}]{parf}
Ying, H.; Jiang, B.; Zhang, J.; Xu, D.; Yu, T.; Dai, Q.; and Fang, L. 2023.
\newblock PARF: Primitive-Aware Radiance Fusion for Indoor Scene Novel View Synthesis.
\newblock In \emph{Proceedings of the IEEE/CVF International Conference on Computer Vision}, 17706--17716.

\bibitem[{Yu et~al.(2021)Yu, Li, Tancik, Li, Ng, and Kanazawa}]{plenoctrees}
Yu, A.; Li, R.; Tancik, M.; Li, H.; Ng, R.; and Kanazawa, A. 2021.
\newblock Plenoctrees for real-time rendering of neural radiance fields.
\newblock In \emph{Proceedings of the IEEE/CVF International Conference on Computer Vision}, 5752--5761.

\bibitem[{Yu et~al.(2022)Yu, Peng, Niemeyer, Sattler, and Geiger}]{monosdf}
Yu, Z.; Peng, S.; Niemeyer, M.; Sattler, T.; and Geiger, A. 2022.
\newblock Monosdf: Exploring monocular geometric cues for neural implicit surface reconstruction.
\newblock \emph{Advances in neural information processing systems}, 35: 25018--25032.

\bibitem[{Zhang et~al.(2023{\natexlab{a}})Zhang, Xing, Zeng, Liu, Shi, and Han}]{wenyuan}
Zhang, W.; Xing, R.; Zeng, Y.; Liu, Y.-S.; Shi, K.; and Han, Z. 2023{\natexlab{a}}.
\newblock Fast Learning Radiance Fields by Shooting Much Fewer Rays.
\newblock \emph{IEEE Transactions on Image Processing}, 32: 2703--2718.

\bibitem[{Zhang et~al.(2023{\natexlab{b}})Zhang, Kundu, Funkhouser, Guibas, Su, and Genova}]{nerflets}
Zhang, X.; Kundu, A.; Funkhouser, T.; Guibas, L.; Su, H.; and Genova, K. 2023{\natexlab{b}}.
\newblock Nerflets: Local radiance fields for efficient structure-aware 3d scene representation from 2d supervision.
\newblock In \emph{Proceedings of the IEEE/CVF Conference on Computer Vision and Pattern Recognition}, 8274--8284.

\end{thebibliography}
\clearpage


\section*{Supplementary material}

# Supplementary Material for *Sparis: Neural Implicit Surface Reconstruction of Indoor Scenes from Sparse Views*

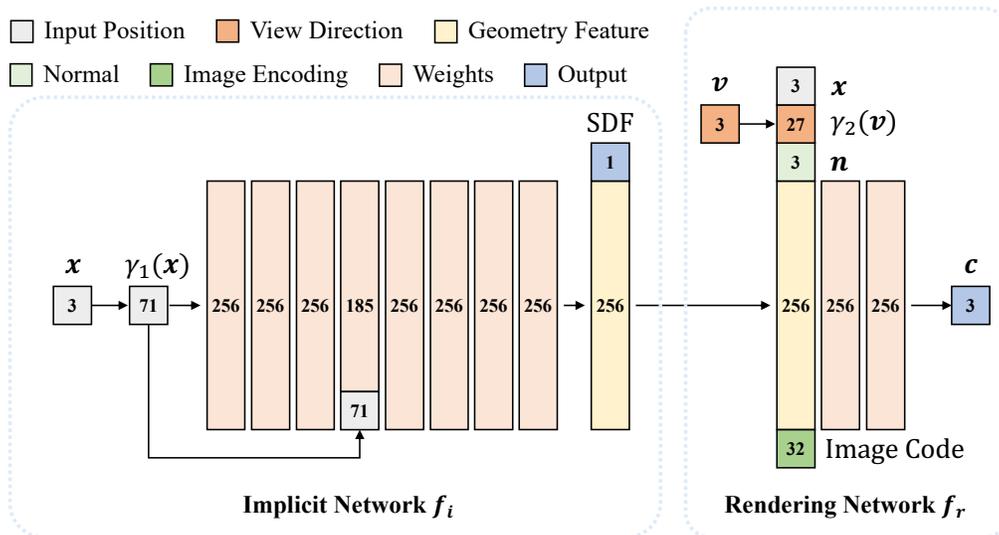

Figure 1: Network architecture of our model. $\gamma_1(\cdot), \gamma_2(\cdot)$ denote Positional Encoding and View Direction Encoding respectively. The image code is encoded from the sample batch using a linear layer. The surface normal $\boldsymbol{n} = \nabla f_i(\boldsymbol{x})$.

In this supplementary document, we provide additional architectural and implementation details in Section **A**. Subsequently, we undertake additional ablation studies on the depth prior to further investigate its impact in Section **B**. Following that, in order to demonstrate the superiority of our method, we conducted experimental comparisons with the SOTA reconstruction method on general objects and the currently popular Gaussian splatting-based approaches, as detailed in Section **C**. In Section **D**, we discuss the sparse pose setting and the fairness of the experiments, and we also provide some insights for future work. Then, we discuss the potential social negative impact of this work in Section **E**. Finally, we present a comprehensive comparison of our experiments in Section **F**.

## A. Methodology and Implementation Details

**Network Architecture**
The detailed network architecture of *Sparis* is illustrated in Figure 1. We use 2 MLPs, an implicit geometry network $f_i$ and a color rendering network $f_r$, to represent the implicit signed distance field and radiance field respectively. The implicit network uses Softplus function

$$\text{Softplus}(\boldsymbol{x}) = \frac{1}{\beta}\log(1+e^{\beta \boldsymbol{x}}) \qquad (1)$$

as activation functions, with $\beta$ set as 100.

**More Experimental Settings**
We use Adam optimizer (Kingma and Ba 2014) with a learning rate of 5e-4 and train the network with batches of 1024 rays for 200k iterations. $\lambda_1, \lambda_2, \lambda_3, \lambda_4, \gamma, \epsilon$ are set to 0.01, 0.01, 0.05, 0.05, 0.1, 0.001. The optimization can be completed in about 12 hours for each scene. The experiments were conducted using an NVIDIA RTX 3090 GPU and two Intel Xeon C6226R CPUs. The software environment was based on Ubuntu 18.04, with PyTorch 1.10.0.

**Evaluation Metrics**
For quantitative comparisons, we employed evaluation metrics from NeuRIS (Wang et al. 2022), including F-score, accuracy, completeness, precision and recall. The definitions of these metrics are shown in Table 1.

## B. More Ablation Study and Comparisons

In the main text, we have explored the influence of $\mathcal{L}_{mono\ depth}$ in MonoSDF (Yu et al. 2022) under sparse view inputs. Following the method in MonoSDF, scaling the monocular depth provided by Ominidata (Eftekhar et al. 2021) during the rendering process fails to deliver effective geometric information.

Given that our method can compute partially relatively accurate absolute depths through inter-image priors and camera poses, a very straightforward idea is to utilize these cal-

| Metric | Definition |
|---|---|
| Accuracy | $\text{mean}_{\boldsymbol{p}\in\mathcal{P}}\left(\min_{\boldsymbol{p}^*\in\mathcal{P}^*}\|\boldsymbol{p}-\boldsymbol{p}^*\|\right)$ |
| Completeness | $\text{mean}_{\boldsymbol{p}^*\in\mathcal{P}^*}\left(\min_{\boldsymbol{p}\in\mathcal{P}}\|\boldsymbol{p}-\boldsymbol{p}^*\|\right)$ |
| Precision | $\text{mean}_{\boldsymbol{p}\in\mathcal{P}}(\min_{\boldsymbol{p}^*\in\mathcal{P}^*}\|\boldsymbol{p}-\boldsymbol{p}^*\| < 0.05)$ |
| Recall | $\text{mean}_{\boldsymbol{p}^*\in\mathcal{P}^*}(\min_{\boldsymbol{p}\in\mathcal{P}}\|\boldsymbol{p}-\boldsymbol{p}^*\| < 0.05)$ |
| F-score | $\frac{2\times\text{Precision}\times\text{Recall}}{\text{Precision}+\text{Recall}}$ |

Table 1: Evaluation metrics in our work. $\mathcal{P}$ and $\mathcal{P}^*$ respectively refer to the point clouds sampled from the surfaces of the predicted mesh and the ground truth mesh.

| $\mathcal{L}_n$ | $\mathcal{L}_{scale\ depth}$ | $\mathcal{L}_{depth}$ | F-score↑ | Acc.↓ | Comp.↓ | Prec.↑ | Recall↑ |
|---|---|---|---|---|---|---|---|
| √ | | | 0.244 | 0.177 | 0.319 | 0.308 | 0.212 |
| √ | √ | | 0.248 | 0.234 | 0.295 | 0.272 | 0.237 |
| √ | | √ | **0.598** | **0.061** | **0.074** | **0.624** | **0.577** |

Table 2: Ablation studies of depth loss in our method over 10 scenes of ScanNet.

culated partial absolute depths to scale the monocular depth from Ominidata, serving as supervision for absolute depth.

To further explore various types of depth supervision, we propose $\mathcal{L}_{scale\ depth}$, which can be mathematically expressed as follows:

$$\mathcal{L}_{scale\ depth} = \sum_{\boldsymbol{r}\in\mathcal{R}}\left\|(\tilde{w}\hat{D}(\boldsymbol{r}) + \tilde{q}) - \bar{D}(\boldsymbol{r})\right\|^2, \quad (2)$$

where $\hat{D}(\boldsymbol{r})$ denotes the rendered depth and $\bar{D}(\boldsymbol{r})$ indicates the monocular depth. $\tilde{w}$ and $\tilde{q}$ are the scale and shift calculated using the least-squares criterion with the inter-image triangulated depth $\widetilde{D}(\boldsymbol{r})$:

$$\tilde{w}, \tilde{q} = \arg\min_{w,q}\sum_{i=1}^{H}\left(w\widetilde{D}(\boldsymbol{r}^i) + q - \bar{D}(\boldsymbol{r}^i)\right)^2. \quad (3)$$

$H$ indicates the number of matching pixel pairs.

The experimental results presented in Table 2 demonstrate that the $\mathcal{L}_{depth}$ proposed in our approach significantly outperforms $\mathcal{L}_{scale\ depth}$. This phenomenon is attributed to reasons discussed in DäRF (Song et al. 2023), where it is mentioned that monocular depth exhibits confusion regarding the scale of different objects. Consequently, employing a strategy that scales the entire image based on absolute depth remains inappropriate. Moreover, without information on different objects within the image and their absolute depths, we are unable to employ a localized scaling approach to improve the results.

## C. Additional Experiments Results
### Experiments on General Object Dataset
To demonstrate the effectiveness of our method, we conducted experiments on a general object dataset following the settings of NeuSurf (Huang et al. 2024b). Table 3 shows that our method has comparable performance to NeuSurf on DTU dataset (Jensen et al. 2014). Ours outperforms the SOTA method NeuSurf in 7 out of 15 DTU scenes.

| Method | Mean CD↓ |
|---|---|
| SparseNeuS (Long et al. 2022) | 3.34 |
| MonoSDF (Yu et al. 2022) | 1.86 |
| NeuSurf (Huang et al. 2024b) | **1.35** |
| Ours | 1.37 |

Table 3: Chamfer Distance on DTU dataset with 3 views.

### Comparison with Gaussian Splitting-based Approaches
SuGaR (Guédon and Lepetit 2023) is a multi-view surface reconstruction method that leverages the Gaussian Splatting pipeline, efficiently achieving surface optimization during the rendering process by binding Gaussians to the mesh surface. DN-Splatter (Turkulainen et al. 2025) regularizes the optimization procedure with depth information, enforces local smoothness of nearby Gaussians, and uses the geometry of the 3D Gaussians supervised by normal cues to achieve better alignment with the true scene geometry. 2DGS (Huang et al. 2024a), as a recent rapid Gaussian splatting surface reconstruction method, flattens the Gaussian primitives from three dimensions to two dimensions, allowing for better coverage of objects and improved geometric learning. Finally, the surface is reconstructed through depth fusion. To verify the effectiveness of our method, we conducted sparse view reconstruction using SuGaR, DN-Splatter and 2DGS on the same dataset, and compared their results with ours, as shown in Table 4.

| Method | F-score↑ | Acc.↓ | Comp.↓ | Prec.↑ | Recal.↑ |
|---|---|---|---|---|---|
| SuGaR (Guédon and Lepetit 2023) | 0.185 | 0.273 | 0.277 | 0.233 | 0.157 |
| DN-Splatter (Turkulainen et al. 2025) | 0.044 | 0.240 | 1.884 | 0.217 | 0.027 |
| 2DGS (Huang et al. 2024a) | 0.137 | 0.215 | 0.512 | 0.291 | 0.090 |
| DUSt3R† (Wang et al. 2024) | 0.565 | 0.059 | 0.088 | 0.612 | 0.526 |
| Ours | **0.647** | **0.056** | **0.060** | **0.666** | **0.631** |

Table 4: Quantitative comparisons of room-scale surface reconstruction results over 10 scenes of ScanNet with 15-20 input views. †GT poses are included as inputs for DUSt3R reconstruction.

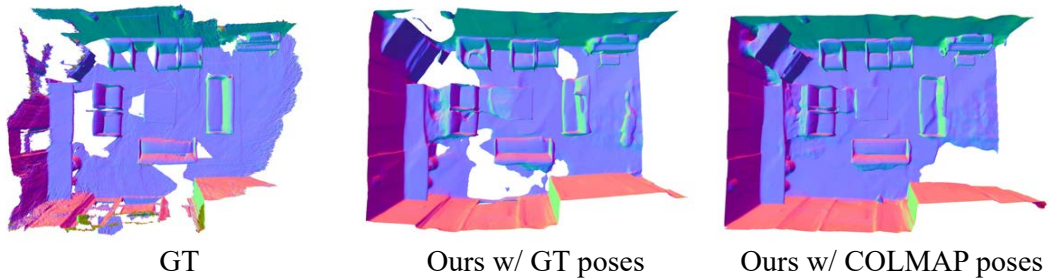

GT      Ours w/ GT poses      Ours w/ COLMAP poses

Figure 2: Visual comparison of reconstruction results with different pose settings on ScanNet with sparse input views.

The results indicate that our approach significantly outperforms the three methods. Gaussian splatting-based reconstruction methods still fall short of achieving the desired outcomes in sparse view reconstruction of indoor scenes. Nevertheless, Gaussian Splatting is a rapidly evolving field, and enhancing the reconstruction quality of such methods under different conditions would present an interesting task.

**Comparison with Large-Scale Pre-Trained Model**

DUSt3R (Wang et al. 2024) is the first holistic end-to-end 3D reconstruction pipeline from un-calibrated and un-posed images. We conducted sparse view reconstruction using it on the same dataset. The quantitative results are presented in 4. Our method excels in sparse conditions and provides a smooth and complete reconstruction.

## D. Discussion and Future Works

**Assumption of View Poses**

It is important to note that the use of ground truth (GT) poses in our method is not due to difficulties in pose estimation. In fact, relatively accurate poses can be obtained using the basic COLMAP tool, as demonstrated in Figure 3. This is feasible because our sparse view setting is consistent with the approach used in DDP-NeRF (Roessle et al. 2022). Therefore, the assumption that view poses are readily available is practically reasonable.

**Use of Ground Truth vs. COLMAP poses**

Our method is primarily designed to reconstruct more accurate surface geometry. To ensure a fair comparison of the mesh quality across different methods, we use ground truth

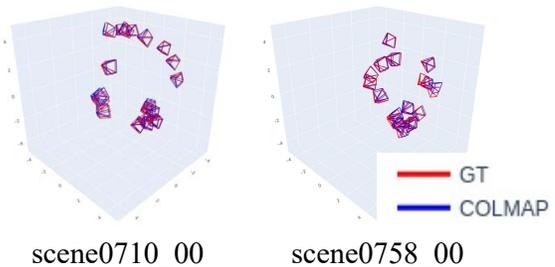

scene0710_00      scene0758_00

Figure 3: Visual comparisons of GT poses and COLMAP poses with sparse views on ScanNet.

| Method | Pose | F-score↑ |
|---|---|---|
| MonoSDF (Yu et al. 2022) | GT | 0.328 |
| NeuRIS (Wang et al. 2022) | GT | 0.464 |
| Ours | COLMAP | 0.514 |
| Ours | GT | **0.647** |

Table 5: Reconstruction quality of ours with different pose settings on 10 ScanNet scenes with sparse views.

(GT) poses aligned with the fused mesh, which helps to mitigate the effects of camera noise. Moreover, methods based on Grid Encoding, such as HelixSurf (Liang et al. 2023) and MonoSDF (w/ Fea. Grids), are known to be sensitive to pose noise, making it necessary to avoid introducing such variations. Consistent with previous works, many general object-

| Method | room0 | | | | | room1 | | | | |
|---|---|---|---|---|---|---|---|---|---|---|
| | F-score↑ | Acc.↓ | Comp.↓ | Prec.↑ | Recal.↑ | F-score↑ | Acc.↓ | Comp.↓ | Prec.↑ | Recal.↑ |
| HelixSurf | 0.043 | 0.555 | 1.035 | 0.056 | 0.035 | 0.031 | 0.489 | 0.842 | 0.041 | 0.024 |
| S$^3$P | 0.006 | 0.351 | 3.551 | 0.091 | 0.003 | 0.015 | 0.154 | 3.113 | 0.190 | 0.008 |
| MonoSDF | 0.526 | 0.055 | 0.233 | 0.671 | 0.433 | 0.620 | 0.057 | 0.062 | 0.630 | 0.611 |
| NeuRIS | 0.527 | 0.054 | 0.202 | 0.595 | 0.472 | 0.421 | 0.076 | 0.087 | 0.449 | 0.396 |
| Ours | 0.800 | 0.039 | 0.111 | 0.852 | 0.753 | 0.903 | 0.024 | 0.038 | 0.940 | 0.868 |

| Method | room2 | | | | | office0 | | | | |
|---|---|---|---|---|---|---|---|---|---|---|
| | F-score↑ | Acc.↓ | Comp.↓ | Prec.↑ | Recal.↑ | F-score↑ | Acc.↓ | Comp.↓ | Prec.↑ | Recal.↑ |
| HelixSurf | 0.014 | 0.500 | 0.900 | 0.020 | 0.010 | 0.005 | 0.667 | 0.802 | 0.006 | 0.004 |
| S$^3$P | 0.024 | 0.350 | 1.655 | 0.104 | 0.013 | 0.044 | 0.397 | 1.162 | 0.128 | 0.026 |
| MonoSDF | 0.367 | 0.073 | 0.139 | 0.410 | 0.332 | 0.440 | 0.091 | 0.195 | 0.497 | 0.394 |
| NeuRIS | 0.433 | 0.081 | 0.127 | 0.458 | 0.410 | 0.365 | 0.082 | 0.179 | 0.441 | 0.311 |
| Ours | 0.833 | 0.028 | 0.085 | 0.892 | 0.782 | 0.813 | 0.028 | 0.075 | 0.871 | 0.763 |

| Method | office1 | | | | | office2 | | | | |
|---|---|---|---|---|---|---|---|---|---|---|
| | F-score↑ | Acc.↓ | Comp.↓ | Prec.↑ | Recal.↑ | F-score↑ | Acc.↓ | Comp.↓ | Prec.↑ | Recal.↑ |
| HelixSurf | - | 0.687 | 1.064 | 0.000 | 0.000 | 0.041 | 0.482 | 0.926 | 0.052 | 0.034 |
| S$^3$P | 0.035 | 0.308 | 1.421 | 0.151 | 0.019 | 0.020 | 0.382 | 1.995 | 0.077 | 0.011 |
| MonoSDF | 0.416 | 0.097 | 0.109 | 0.413 | 0.419 | 0.479 | 0.100 | 0.103 | 0.488 | 0.470 |
| NeuRIS | 0.351 | 0.088 | 0.152 | 0.438 | 0.293 | 0.380 | 0.086 | 0.136 | 0.415 | 0.351 |
| Ours | 0.760 | 0.027 | 0.084 | 0.870 | 0.675 | 0.842 | 0.033 | 0.038 | 0.855 | 0.829 |

| Method | office3 | | | | | office4 | | | | |
|---|---|---|---|---|---|---|---|---|---|---|
| | F-score↑ | Acc.↓ | Comp.↓ | Prec.↑ | Recal.↑ | F-score↑ | Acc.↓ | Comp.↓ | Prec.↑ | Recal.↑ |
| HelixSurf | 0.026 | 0.528 | 0.988 | 0.037 | 0.020 | 0.039 | 0.556 | 0.859 | 0.065 | 0.028 |
| S$^3$P | 0.001 | 0.095 | 4.837 | 0.247 | 0.001 | 0.004 | 0.133 | 4.126 | 0.227 | 0.002 |
| MonoSDF | 0.482 | 0.079 | 0.128 | 0.530 | 0.442 | 0.305 | 0.098 | 0.144 | 0.335 | 0.279 |
| NeuRIS | 0.451 | 0.069 | 0.118 | 0.498 | 0.412 | 0.520 | 0.054 | 0.174 | 0.618 | 0.449 |
| Ours | 0.823 | 0.034 | 0.067 | 0.877 | 0.775 | 0.823 | 0.033 | 0.088 | 0.888 | 0.767 |

Table 6: Quantitative comparisons of surface reconstruction results on individual scenes of Replica with 10 input views.

level approaches (e.g., SparseNeuS, NeuSurf) also utilize GT poses as input. However, to illustrate the robustness of our method against pose noise, we applied our approach to all 10 ScanNet scenes using COLMAP poses obtained from sparse views. Importantly, this does not alter our sparse setting but serves to provide illustrative results. As shown in Table 5 and Figure 2, even when using poses estimated from sparse images, our method still produces superior geometry compared to other methods that rely on GT poses.

**Future works**

Although we fully leverage single-view priors and inter-image priors, the reconstruction results still suffer from imperfections due to incomplete viewpoint coverage. In the future, we will explore how to incorporate additional information to fill in areas not covered by sparse views; the diffusion model might be an interesting idea to consider.

## E. Potential Negative Social Impacts

Our method for indoor reconstruction, which necessitates only a dozen or so collected images, may more easily lead to unauthorized reconstruction of private houses, thus infringing on privacy issues, due to its reliance on sparse images. Our reconstruction process requires the use of high-performance GPUs for several hours of computing, which may increase energy consumption and carbon emissions.

## F. More Quantitative and Visual Comparisons

We provide comprehensive quantitative and visual comparisons of room-scale surface reconstruction results. Table 6 presents detailed quantitative evaluations for individual scenes from the Replica dataset using 10 input views. Similarly, Table 7 offers quantitative comparisons for individual scenes from the ScanNet dataset with 15-20 input views.

For visual assessments, Figure 4 and Figure 5 showcase the reconstruction results on various scenes from the ScanNet dataset, highlighting differences across scenes with sparse input views. Additionally, Figure 6 illustrates visual comparisons for multiple scenes from the Replica dataset, including different rooms and offices, also using sparse input views. These comparisons provide a thorough examination of the performance across different datasets and scenarios.

| Method | scene0009_01 | | | | | scene0050_00 | | | | |
|---|---|---|---|---|---|---|---|---|---|---|
| | F-score↑ | Acc.↓ | Comp.↓ | Prec.↑ | Recal.↑ | F-score↑ | Acc.↓ | Comp.↓ | Prec.↑ | Recal.↑ |
| COLMAP | 0.131 | 0.414 | 0.722 | 0.178 | 0.104 | 0.082 | 0.102 | 0.898 | 0.350 | 0.047 |
| DDP-NeRF | 0.181 | 0.456 | 0.104 | 0.112 | 0.479 | 0.408 | 0.140 | 0.079 | 0.302 | 0.629 |
| DäRF | 0.269 | 0.473 | 0.091 | 0.185 | 0.489 | 0.259 | 0.302 | 0.098 | 0.189 | 0.412 |
| NeuS | 0.193 | 0.181 | 0.330 | 0.240 | 0.162 | - | - | - | - | - |
| VolRecon | 0.156 | 0.209 | 0.231 | 0.168 | 0.147 | 0.149 | 0.161 | 0.334 | 0.205 | 0.118 |
| HelixSurf | 0.261 | 0.173 | 0.228 | 0.276 | 0.248 | 0.481 | 0.098 | 0.090 | 0.497 | 0.466 |
| S$^3$P | 0.306 | 0.272 | 0.160 | 0.280 | 0.337 | 0.303 | 0.188 | 0.151 | 0.324 | 0.285 |
| MonoSDF | 0.348 | 0.119 | 0.155 | 0.349 | 0.347 | 0.042 | 0.305 | 0.204 | 0.034 | 0.056 |
| NeuRIS | 0.649 | 0.066 | 0.045 | 0.633 | 0.667 | 0.217 | 0.437 | 0.159 | 0.220 | 0.213 |
| Ours | 0.725 | 0.044 | 0.065 | 0.764 | 0.690 | 0.702 | 0.042 | 0.056 | 0.744 | 0.664 |

| Method | scene0084_00 | | | | | scene0085_00 | | | | |
|---|---|---|---|---|---|---|---|---|---|---|
| | F-score↑ | Acc.↓ | Comp.↓ | Prec.↑ | Recal.↑ | F-score↑ | Acc.↓ | Comp.↓ | Prec.↑ | Recal.↑ |
| COLMAP | 0.128 | 0.199 | 0.448 | 0.232 | 0.089 | 0.194 | 0.136 | 0.334 | 0.249 | 0.159 |
| DDP-NeRF | 0.420 | 0.155 | 0.063 | 0.317 | 0.621 | 0.234 | 0.240 | 0.078 | 0.150 | 0.535 |
| DäRF | 0.384 | 0.209 | 0.105 | 0.332 | 0.455 | 0.271 | 0.195 | 0.132 | 0.222 | 0.349 |
| NeuS | - | - | - | - | - | - | - | - | - | - |
| VolRecon | 0.139 | 0.210 | 0.268 | 0.177 | 0.114 | 0.141 | 0.268 | 0.242 | 0.141 | 0.141 |
| HelixSurf | 0.056 | 1.110 | 0.322 | 0.050 | 0.063 | 0.301 | 0.175 | 0.250 | 0.333 | 0.275 |
| S$^3$P | 0.129 | 0.652 | 0.254 | 0.102 | 0.175 | 0.326 | 0.274 | 0.172 | 0.322 | 0.331 |
| MonoSDF | 0.075 | 1.555 | 0.382 | 0.063 | 0.094 | 0.601 | 0.060 | 0.063 | 0.609 | 0.594 |
| NeuRIS | 0.311 | 0.274 | 0.106 | 0.285 | 0.342 | 0.659 | 0.053 | 0.055 | 0.663 | 0.655 |
| Ours | 0.763 | 0.040 | 0.041 | 0.775 | 0.751 | 0.643 | 0.047 | 0.076 | 0.684 | 0.607 |

| Method | scene0580_00 | | | | | scene0710_00 | | | | |
|---|---|---|---|---|---|---|---|---|---|---|
| | F-score↑ | Acc.↓ | Comp.↓ | Prec.↑ | Recal.↑ | F-score↑ | Acc.↓ | Comp.↓ | Prec.↑ | Recal.↑ |
| COLMAP | 0.122 | 0.195 | 0.351 | 0.207 | 0.086 | 0.243 | 0.118 | 0.249 | 0.349 | 0.186 |
| DDP-NeRF | 0.190 | 0.332 | 0.061 | 0.115 | 0.562 | 0.221 | 0.326 | 0.073 | 0.141 | 0.517 |
| DäRF | 0.258 | 0.199 | 0.110 | 0.224 | 0.303 | 0.306 | 0.240 | 0.154 | 0.277 | 0.343 |
| NeuS | 0.192 | 0.225 | 0.332 | 0.258 | 0.153 | - | - | - | - | - |
| VolRecon | 0.104 | 0.225 | 0.250 | 0.120 | 0.092 | 0.203 | 0.189 | 0.250 | 0.239 | 0.176 |
| HelixSurf | 0.508 | 0.105 | 0.075 | 0.507 | 0.509 | 0.030 | 0.321 | 0.443 | 0.045 | 0.023 |
| S$^3$P | 0.448 | 0.090 | 0.100 | 0.471 | 0.427 | 0.288 | 0.143 | 0.173 | 0.318 | 0.263 |
| MonoSDF | 0.327 | 0.100 | 0.094 | 0.321 | 0.334 | 0.483 | 0.080 | 0.079 | 0.494 | 0.472 |
| NeuRIS | 0.521 | 0.122 | 0.053 | 0.463 | 0.596 | 0.507 | 0.068 | 0.072 | 0.528 | 0.487 |
| Ours | 0.660 | 0.044 | 0.051 | 0.681 | 0.641 | 0.493 | 0.057 | 0.065 | 0.516 | 0.471 |

| Method | scene0721_00 | | | | | scene0738_00 | | | | |
|---|---|---|---|---|---|---|---|---|---|---|
| | F-score↑ | Acc.↓ | Comp.↓ | Prec.↑ | Recal.↑ | F-score↑ | Acc.↓ | Comp.↓ | Prec.↑ | Recal.↑ |
| COLMAP | 0.040 | 0.087 | 1.815 | 0.423 | 0.021 | 0.257 | 0.107 | 0.286 | 0.343 | 0.205 |
| DDP-NeRF | 0.232 | 0.450 | 0.065 | 0.146 | 0.565 | 0.331 | 0.215 | 0.087 | 0.243 | 0.517 |
| DäRF | 0.236 | 0.421 | 0.155 | 0.176 | 0.357 | 0.258 | 0.279 | 0.157 | 0.199 | 0.364 |
| NeuS | 0.051 | 0.495 | 1.367 | 0.097 | 0.035 | 0.084 | 0.371 | 0.410 | 0.103 | 0.071 |
| VolRecon | 0.056 | 0.295 | 0.520 | 0.102 | 0.039 | 0.276 | 0.186 | 0.183 | 0.268 | 0.284 |
| HelixSurf | - | - | - | - | - | 0.109 | 0.301 | 0.329 | 0.117 | 0.102 |
| S$^3$P | 0.120 | 0.720 | 0.295 | 0.103 | 0.144 | 0.244 | 0.263 | 0.149 | 0.225 | 0.266 |
| MonoSDF | 0.126 | 0.289 | 0.191 | 0.115 | 0.139 | 0.139 | 0.616 | 0.235 | 0.111 | 0.187 |
| NeuRIS | 0.517 | 0.122 | 0.063 | 0.487 | 0.551 | 0.225 | 0.449 | 0.138 | 0.188 | 0.280 |
| Ours | 0.626 | 0.053 | 0.070 | 0.676 | 0.583 | 0.617 | 0.082 | 0.054 | 0.596 | 0.640 |

| Method | scene0758_00 | | | | | scene0781_00 | | | | |
|---|---|---|---|---|---|---|---|---|---|---|
| | F-score↑ | Acc.↓ | Comp.↓ | Prec.↑ | Recal.↑ | F-score↑ | Acc.↓ | Comp.↓ | Prec.↑ | Recal.↑ |
| COLMAP | 0.225 | 0.224 | 0.397 | 0.290 | 0.183 | 0.188 | 0.211 | 0.331 | 0.219 | 0.165 |
| DDP-NeRF | 0.407 | 0.135 | 0.064 | 0.306 | 0.609 | 0.243 | 0.357 | 0.129 | 0.185 | 0.352 |
| DäRF | 0.421 | 0.099 | 0.106 | 0.395 | 0.451 | 0.287 | 0.316 | 0.164 | 0.221 | 0.411 |
| NeuS | 0.211 | 0.173 | 0.346 | 0.285 | 0.168 | 0.062 | 0.358 | 1.205 | 0.128 | 0.041 |
| VolRecon | 0.166 | 0.269 | 0.233 | 0.164 | 0.169 | 0.158 | 0.234 | 0.325 | 0.154 | 0.163 |
| HelixSurf | 0.344 | 0.186 | 0.188 | 0.362 | 0.328 | 0.051 | 0.602 | 0.317 | 0.056 | 0.047 |
| S$^3$P | 0.278 | 0.210 | 0.197 | 0.293 | 0.265 | 0.326 | 0.191 | 0.115 | 0.296 | 0.362 |
| MonoSDF | 0.565 | 0.071 | 0.058 | 0.556 | 0.575 | 0.575 | 0.085 | 0.060 | 0.546 | 0.608 |
| NeuRIS | 0.596 | 0.077 | 0.054 | 0.574 | 0.619 | 0.433 | 0.135 | 0.074 | 0.404 | 0.466 |
| Ours | 0.600 | 0.068 | 0.055 | 0.595 | 0.604 | 0.642 | 0.079 | 0.065 | 0.629 | 0.655 |

Table 7: Quantitative comparisons of reconstruction results on individual scenes of ScanNet with 15-20 input views.

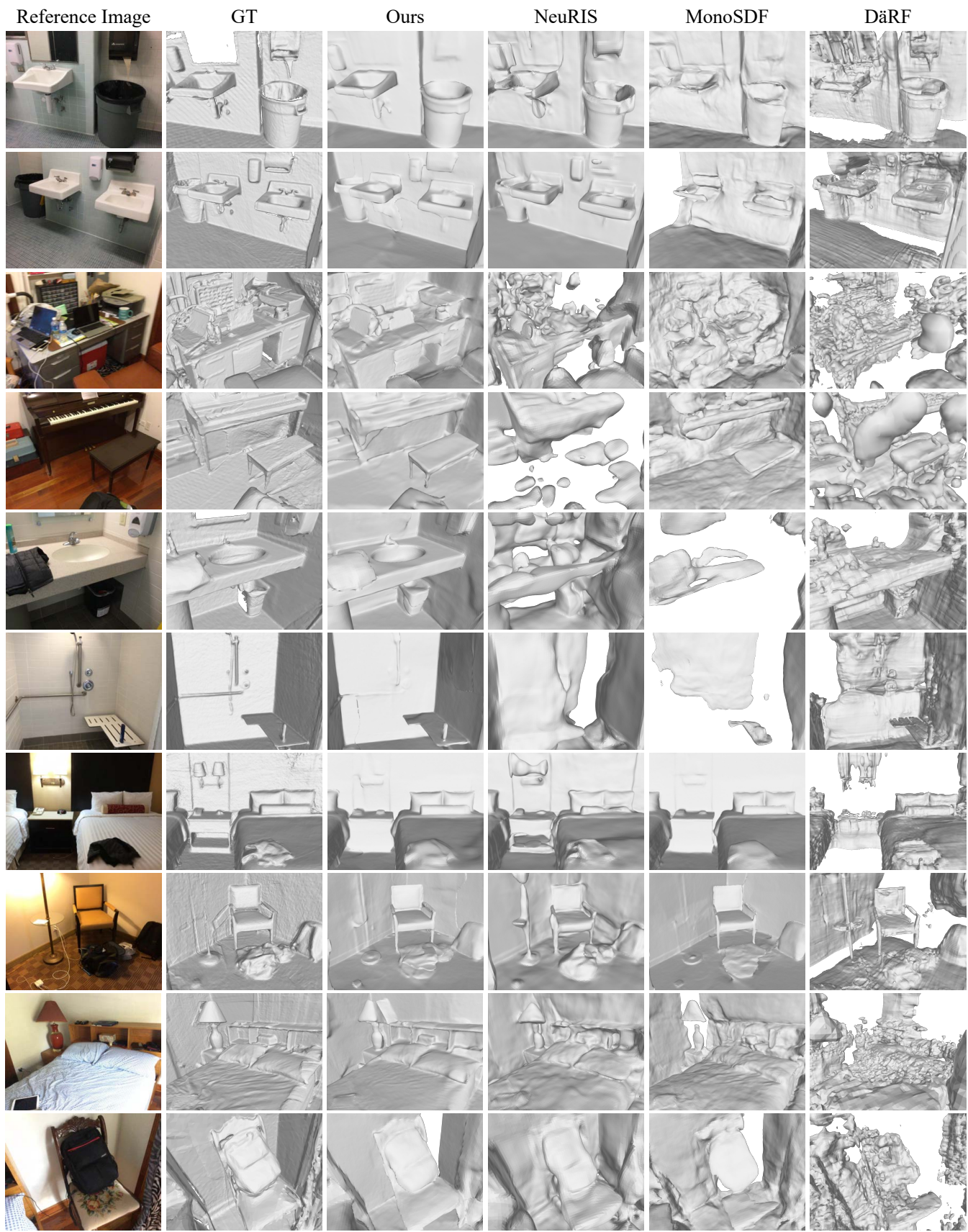

Figure 4: Visual comparisons of room-scale surface reconstruction results on scene0009_01, scene0050_00, scene0084_00, scene0085_00 and scene0580_00 of ScanNet with sparse input views.

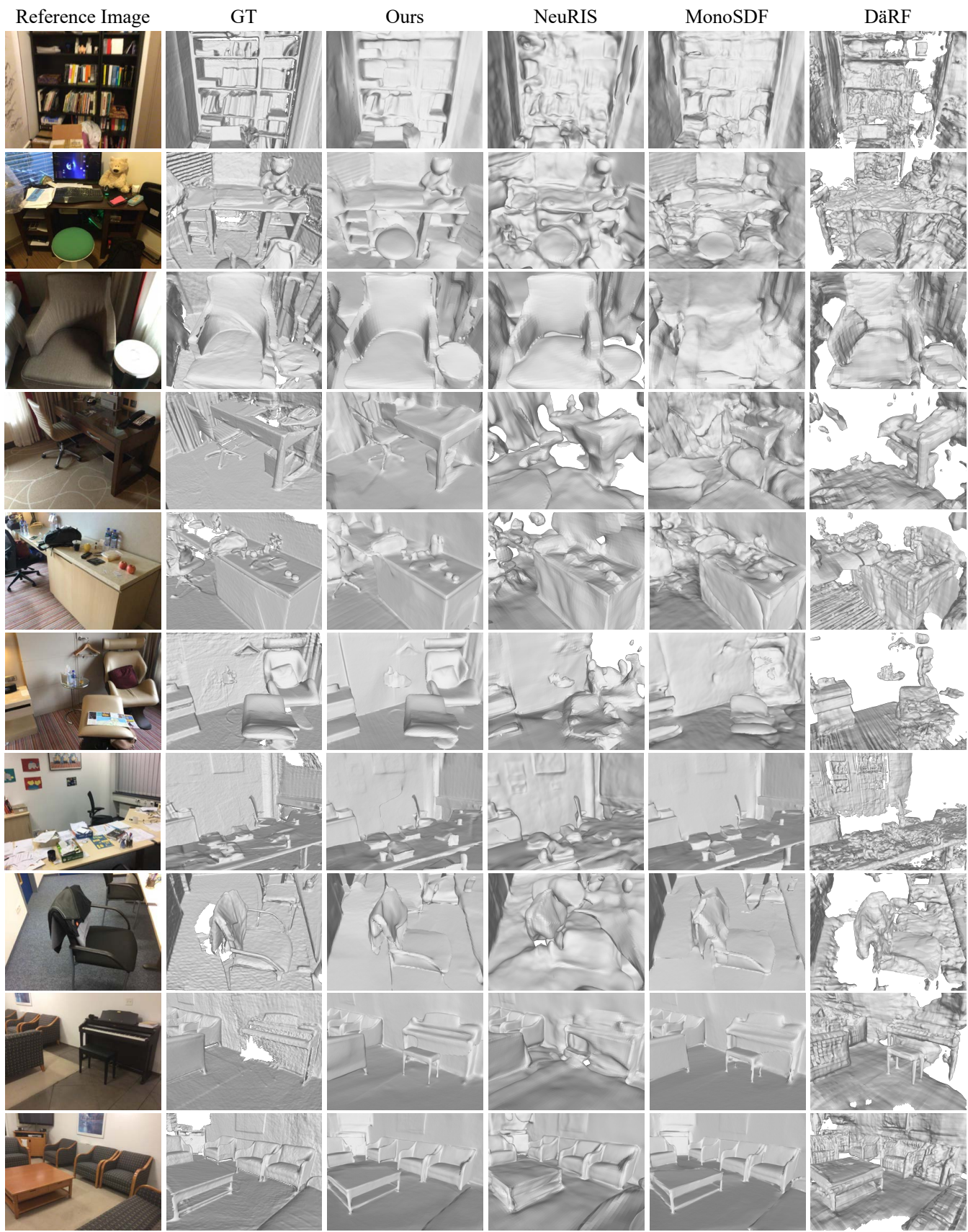

Figure 5: Visual comparisons of room-scale surface reconstruction results on scene0710_00, scene0721_00, scene0738_00, scene0758_00 and scene0781_00 of ScanNet with sparse input views.

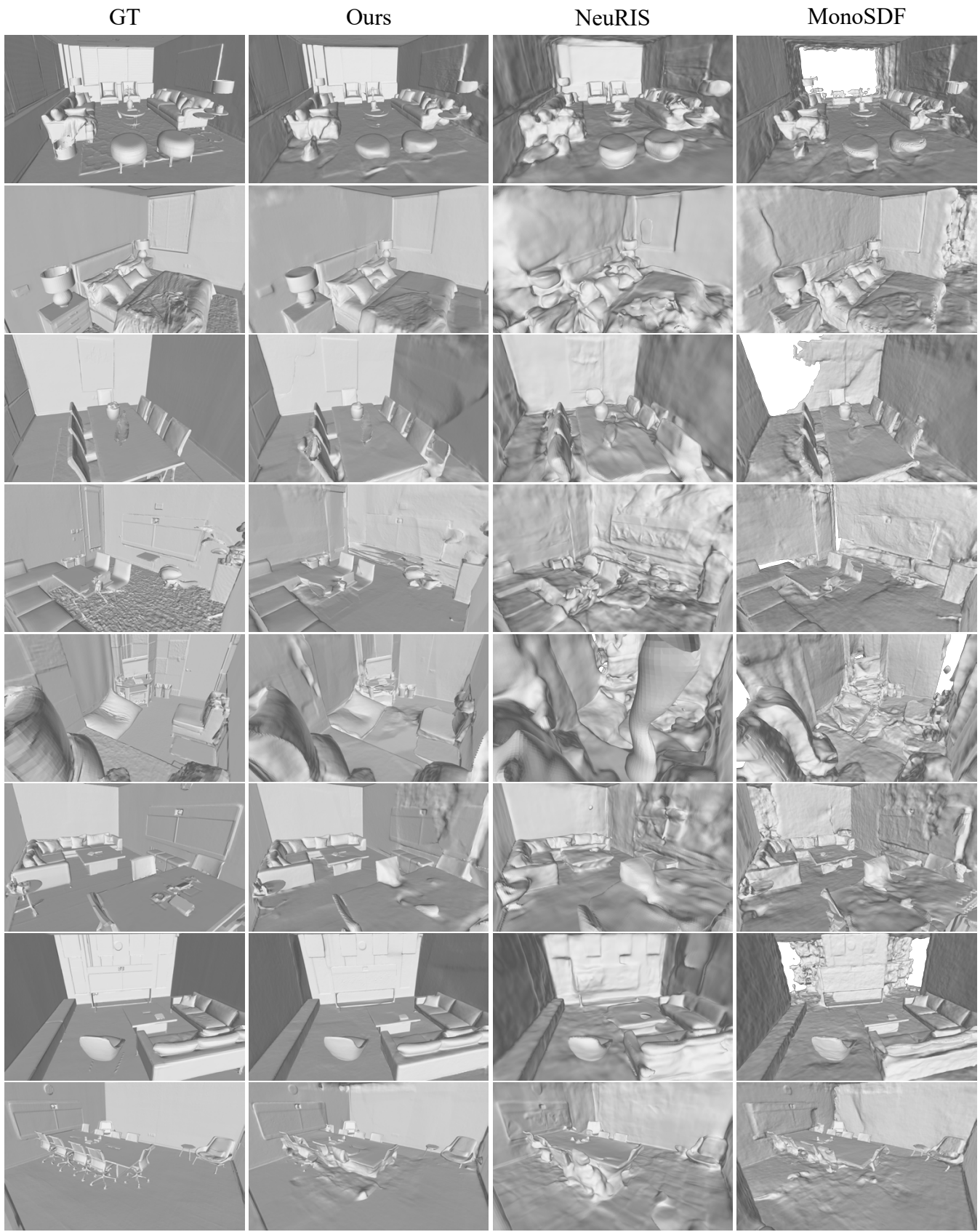

Figure 6: Visual comparisons of room-scale surface reconstruction results on room0, room1, room2, office0, office1, office2, office3 and office4 of Replica with sparse input views.

# References

Eftekhar, A.; Sax, A.; Malik, J.; and Zamir, A. 2021. Omnidata: A Scalable Pipeline for Making Multi-Task Mid-Level Vision Datasets From 3D Scans. In *Proceedings of the IEEE/CVF International Conference on Computer Vision (ICCV)*, 10786–10796.

Guédon, A.; and Lepetit, V. 2023. SuGaR: Surface-Aligned Gaussian Splatting for Efficient 3D Mesh Reconstruction and High-Quality Mesh Rendering. *arXiv preprint arXiv:2311.12775*.

Huang, B.; Yu, Z.; Chen, A.; Geiger, A.; and Gao, S. 2024a. 2d gaussian splatting for geometrically accurate radiance fields. In *ACM SIGGRAPH 2024 Conference Papers*, 1–11.

Huang, H.; Wu, Y.; Zhou, J.; Gao, G.; Gu, M.; and Liu, Y.-S. 2024b. Neusurf: On-surface priors for neural surface reconstruction from sparse input views. In *Proceedings of the AAAI Conference on Artificial Intelligence*, volume 38, 2312–2320.

Jensen, R.; Dahl, A.; Vogiatzis, G.; Tola, E.; and Aanæs, H. 2014. Large scale multi-view stereopsis evaluation. In *2014 IEEE Conference on Computer Vision and Pattern Recognition*, 406–413. IEEE.

Kingma, D. P.; and Ba, J. 2014. Adam: A method for stochastic optimization. *arXiv preprint arXiv:1412.6980*.

Liang, Z.; Huang, Z.; Ding, C.; and Jia, K. 2023. Helixsurf: A robust and efficient neural implicit surface learning of indoor scenes with iterative intertwined regularization. In *Proceedings of the IEEE/CVF Conference on Computer Vision and Pattern Recognition*, 13165–13174.

Long, X.; Lin, C.; Wang, P.; Komura, T.; and Wang, W. 2022. Sparseneus: Fast generalizable neural surface reconstruction from sparse views. In *European Conference on Computer Vision*, 210–227. Springer.

Roessle, B.; Barron, J. T.; Mildenhall, B.; Srinivasan, P. P.; and Nießner, M. 2022. Dense depth priors for neural radiance fields from sparse input views. In *Proceedings of the IEEE/CVF Conference on Computer Vision and Pattern Recognition*, 12892–12901.

Song, J.; Park, S.; An, H.; Cho, S.; Kwak, M.-S.; Cho, S.; and Kim, S. 2023. DäRF: Boosting Radiance Fields from Sparse Inputs with Monocular Depth Adaptation. arXiv:2305.19201.

Turkulainen, M.; Ren, X.; Melekhov, I.; Seiskari, O.; Rahtu, E.; and Kannala, J. 2025. DN-Splatter: Depth and Normal Priors for Gaussian Splatting and Meshing. In *Proceedings of the IEEE/CVF Winter Conference on Applications of Computer Vision (WACV)*.

Wang, J.; Wang, P.; Long, X.; Theobalt, C.; Komura, T.; Liu, L.; and Wang, W. 2022. Neuris: Neural reconstruction of indoor scenes using normal priors. In *European Conference on Computer Vision*, 139–155. Springer.

Wang, S.; Leroy, V.; Cabon, Y.; Chidlovskii, B.; and Revaud, J. 2024. DUSt3R: Geometric 3D Vision Made Easy. In *CVPR*.

Yu, Z.; Peng, S.; Niemeyer, M.; Sattler, T.; and Geiger, A. 2022. Monosdf: Exploring monocular geometric cues for neural implicit surface reconstruction. *Advances in neural information processing systems*, 35: 25018–25032.



\end{document}